\documentclass[10pt]{article}
\usepackage[utf8]{inputenc}
\usepackage[T1]{fontenc}
\usepackage[english]{babel}
\usepackage[numbers]{natbib}
\usepackage{graphicx}
\usepackage[parfill]{parskip}
\usepackage[justification=centering]{caption}
\usepackage{hyperref}
\usepackage{appendix}
\usepackage{subfigure}
\usepackage{placeins}
\usepackage{float}
\usepackage{amsmath}
\usepackage{times}
\usepackage{indentfirst}
\usepackage{multicol}
\usepackage{xcolor}
\usepackage{amsfonts}

\usepackage{pgfplots}
\pgfplotsset{compat=1.18}

\usepackage[utf8]{inputenc}
\usepackage[T1]{fontenc}
\usepackage{booktabs}
\usepackage{array}
\usepackage{xcolor}
\usepackage{multirow}

\usepackage[letterpaper,top=2cm,bottom=2cm,left=3.5cm,right=3.5cm,marginparwidth=1.75cm]{geometry}

\usepackage{graphicx}
\usepackage{array}

\hypersetup{
    colorlinks,
    citecolor=black,
    filecolor=black,
    linkcolor=black,
    urlcolor=black
}

\begin{document}

\captionsetup[figure]{labelfont={bf,it},textfont=it}
\captionsetup[table]{labelfont={bf,it},textfont=it}

\begin{titlepage}
    \centering
    \vspace*{\baselineskip}
    \rule{\textwidth}{1.6pt}\vspace*{-\baselineskip}\vspace*{2pt}
    \rule{\textwidth}{0.4pt}\\[\baselineskip]
    {\LARGE \textbf{Virtual Try-On} \vspace{0,5cm} \\ \Large{Report Stage INF594} \\[0.3\baselineskip]  }
    
    \rule{\textwidth}{0.4pt}\vspace*{-\baselineskip}\vspace{3.2pt}
    \rule{\textwidth}{1.6pt}\\[\baselineskip]
    \scshape
    \vspace*{0.6\baselineskip}
    {\Large Max Rehman Linder \par}
    \vspace*{5\baselineskip}
    {\scshape  This report was submitted August 2024.\par It was edited and published for the public in 2026 }
    {\large}
    \vfill
    {\large Publicis Resources GenAI Team \& École Polytechnique \par}
    max.rehmanlinder@publicisresources.com / max.rehman-linder@polytechnique.edu / maxrl@kth.se
  \end{titlepage}

\section{Abstract}

The field of visual computing has advanced rapidly in the past few years. Notably, latent diffusion models, introduced in 2022 \cite{latent2022}, have been widely adopted, largely due to the release of open-source models. These open-source models have enabled both industry and academia to fine-tune them for various tasks. One of the most widely adopted models is Stable Diffusion, which had its latest open-source version released in 2023 \cite{podell2023sdxlimprovinglatentdiffusion}. While these models can create extremely diverse images, one challenge with latent diffusion models is creating images that adhere closely to very precise conditions. This includes both incorporating specific visual elements and ensuring their accurate placement in the generated image.

In this technical report, I present a new method for guiding image generation in the context of Virtual-Try-On (VITON). The proposed method leverages new open source Ai models to augment the image data with labels, such as lengths and styles. By training adapters with these labels paired with images of the garments, the model can produce a more diverse set of images that the user can control. For the end user, such as a retailer, this means that they can assure that the produced image is as true to the true fit as possible, not misleading consumers. 

\newpage

\tableofcontents

\newpage

\section{Introduction}

VITON can be described as the task of determining how a garment would look on a person, given a picture of the person and the garment. One way of interpreting this task is as an in-painting problem, where the region where the garment would fit is masked, and, given a picture of the garment, we generate an image of the person wearing it. This problem is of significant interest to companies and customers involved in online clothing retail, where trying on clothes can only be done virtually. Other methods for generating similar images also exist, such as IMAGDressing-v1 \cite{shen2024imagdressingv1customizablevirtualdressing}, which involves virtual dressing based on a photo of a person’s face to generate an image of their entire body. However, this report is limited in scope to VITON.

Extensive work has been done in this area over the past few years using various methods. Early approaches, such as GANs like Viton-GAN \cite{GAN} and VitonHD \cite{choi2021vitonhdhighresolutionvirtualtryon}, were not very powerful. The inherent weakness of purpose-trained GANs is that they lack the generalist ability to produce photorealistic images like diffusion models do. These models attempt to warp the garment to fit onto the masked area, but for believable images, lighting, folds, and other details must also be generated. Additionally, if the masked area extends beyond the garment (such as covering arms or legs), these models struggle to generate those parts realistically.

This is where diffusion models excel. Pre-trained diffusion models have the inherent capability to generate photorealistic images. However, pre-trained text-to-image (T2I) models lack the ability to precisely place objects in an image, such as details on a shirt. To address this, methods like T2I-adapters \cite{mou2023t2iadapterlearningadaptersdig} and ControlNets \cite{zhang2023addingconditionalcontroltexttoimage} have been developed, providing meticulous control over contours and placements. However, these techniques cannot be directly applied to VITON because, unlike other tasks, we do not know exactly where different details of the garment will end up on the person, as this depends on the person’s pose and the garment’s fit. This requires a learnable transition.

For this task, the authors of Animate Anyone \cite{hu2024animateanyoneconsistentcontrollable} developed a process using parallel U-Nets with shared self-attention, where features are shared between one U-Net and another to generate video from a photo. Similar to VITON, maintaining coherence between images/frames while allowing details to change positions is crucial. This method was first implemented by IDM-VITON \cite{choi2024improvingdiffusionmodelsauthentic} in May 2024 for this purpose.

Every diffusion model developed for this purpose has used various U-Nets in some form, where information is passed from a static U-Net to a denoising one. This includes the first ever published model, Stable VITON \cite{kim2023stablevitonlearningsemanticcorrespondence}, Google's closed-source TryOnDiffusion \cite{zhu2023tryondiffusiontaleunets}, and very recent models like IDM-VITON \cite{choi2024improvingdiffusionmodelsauthentic}. These models generate photorealistic images. However, they lack the ability to meaningfully control the fit and style of the garment. For instance, clothing vendors who want their garments to fit the image as they do in real life would require control over the fit. Similarly, typical users of a VITON model might want to style a garment as they would in real life for authenticity.

In this report, I explore the possibility of conditioning the model to provide more control over how the garment is outfitted. This is achieved by adding extra labels to the dataset of images used, as well as masking the image in various ways and using that as a condition. Semantic labels were created using GPT-4, while more detailed labels about the position of the garment were created using custom methods. These labels—a mix of one-hot encoded classes and floats describing the exact position of garments—are concatenated into a single vector and projected into an IP-adapter using a custom-made module named ControlableClothing, which can be added to the standard IP-adapter.

\section{Related works}

This section aims to look at methods used for my work as well as other related works withing the field of VITON.

\subsection{Diffusion Models}
\begin{figure}[h]
    \centering
    \begin{minipage}[b]{0.24\textwidth}
        \centering
        \includegraphics[width=\textwidth]{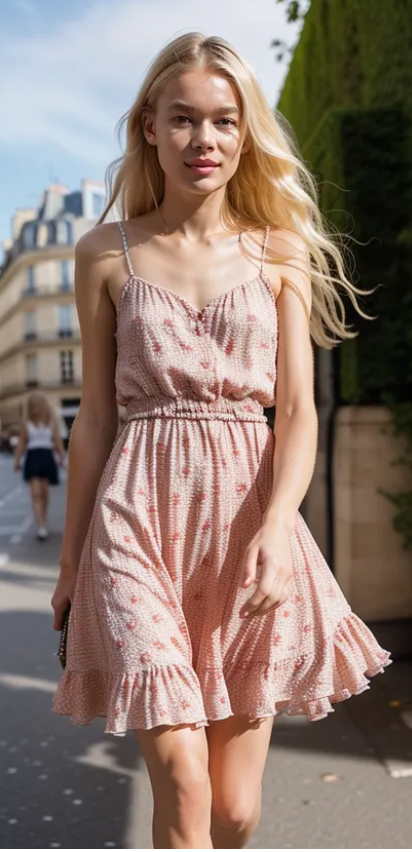}
        \caption*{Original}
    \end{minipage}
    \hfill
    \begin{minipage}[b]{0.24\textwidth}
        \centering
        \includegraphics[width=\textwidth]{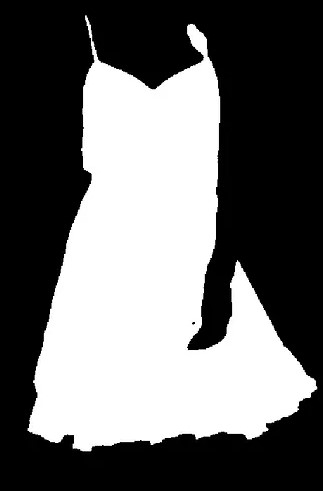}
        \caption*{Masked}
    \end{minipage}
    \hfill
    \begin{minipage}[b]{0.24\textwidth}
        \centering
        \includegraphics[width=\textwidth]{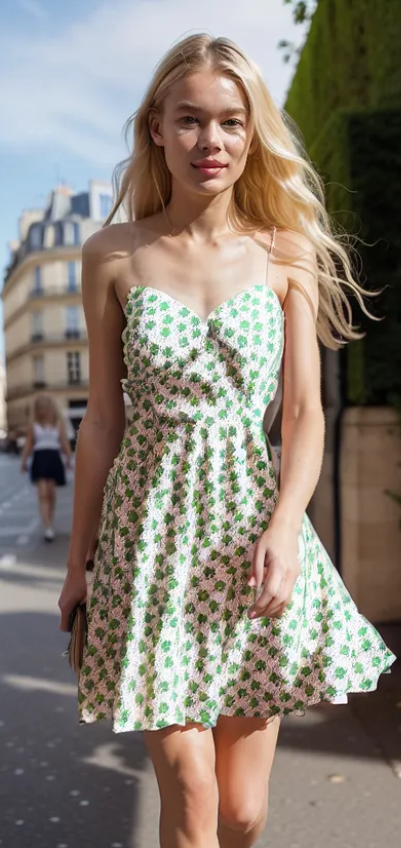}
        \caption*{"green \& white"}
    \end{minipage}
    \hfill
    \begin{minipage}[b]{0.24\textwidth}
        \centering
        \includegraphics[width=\textwidth]{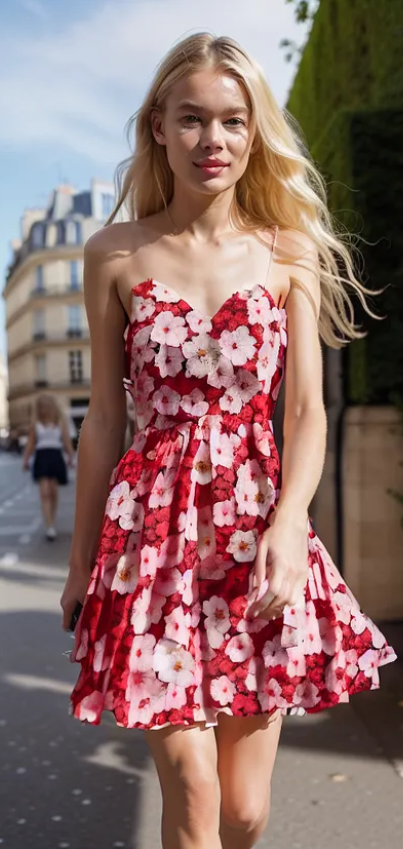}
        \caption*{"Red Flowers"}
    \end{minipage}
    \caption{How in-painting can be used to create photo realistic images}
    \label{fig:in-paint}
\end{figure}

At the core of the model, is Stable Diffusion SDXL \cite{podell2023sdxlimprovinglatentdiffusion}, which was released in July 2023 and remains powerful T2I-model as of 2024. It has become the most widely adopted model of its kind by the open-source community and has many useful extra modules and variants. Among which is the IP-adapter \cite{ye2023ipadaptertextcompatibleimage}, the SDXL inpainitng-pipeline and SDXL image-to-image. \\
\\
Starting with SDXL-inpainting, it is already able to in-paint clothes and other objects via prompt. However, the control over the in-paint is limited to how we mask the image and the prompt we can come up with, as seen in Figure \ref{fig:in-paint}. For the purpose of generating fun images,this is sufficient bu not for VITON where no prompt can capture every detail of a garment.

\subsection{Diffusion models for Virtual Try-On}

\subsubsection{TryOnDiffusion}

The first published paper on diffusion models for virtual try-on published summer of 2023 (to my knowledge), is TryOnDiffusion by Google \cite{radford2021learningtransferablevisualmodels}. 

\begin{figure}[H]
    \centering
    \includegraphics[width=0.7\textwidth]{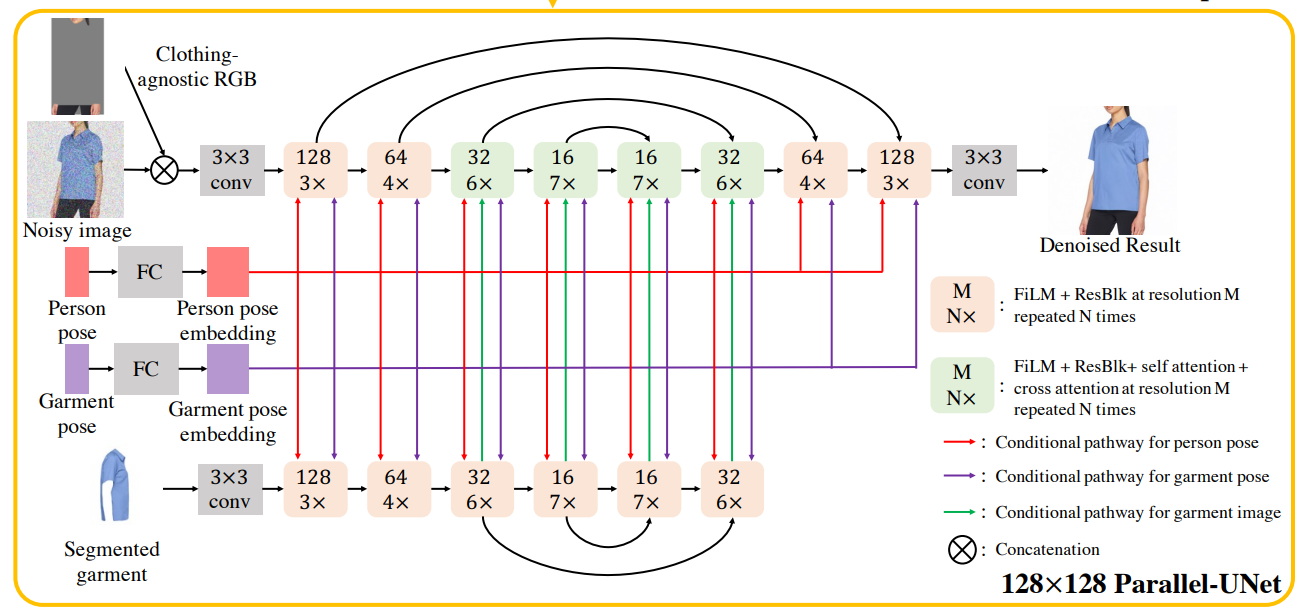}
    \caption{The architecture of TryOnDiffusion uses parallel unets in the diffusion process.}
    \label{TryOnDiffusion}
\end{figure}
This model were the first that showed reliably "good" results, and compared to all other models that it was compared to (all GANs at the time), it greatly outperformed everyone of them. This model was never released publicly and the massive dataset of 4 millions images were scraped of the internet. So whereas this model could never be tested independently or used commercially by google, it made clear that diffusion models could outperform all GANs.

\subsubsection{OOTD Diffusion and IDM-viton}

OOTD-diffusion \cite{xu2024ootdiffusionoutfittingfusionbased} was released in match of 2024 and utilizes parallel U-nets from stable diffusion. It works by having the tensors that goes into self-attention concatenated. This method was also used in IDM-viton \cite{choi2024improvingdiffusionmodelsauthentic}, but with greater success.

\begin{figure}[H]
    \centering
    \includegraphics[width=0.8\textwidth]{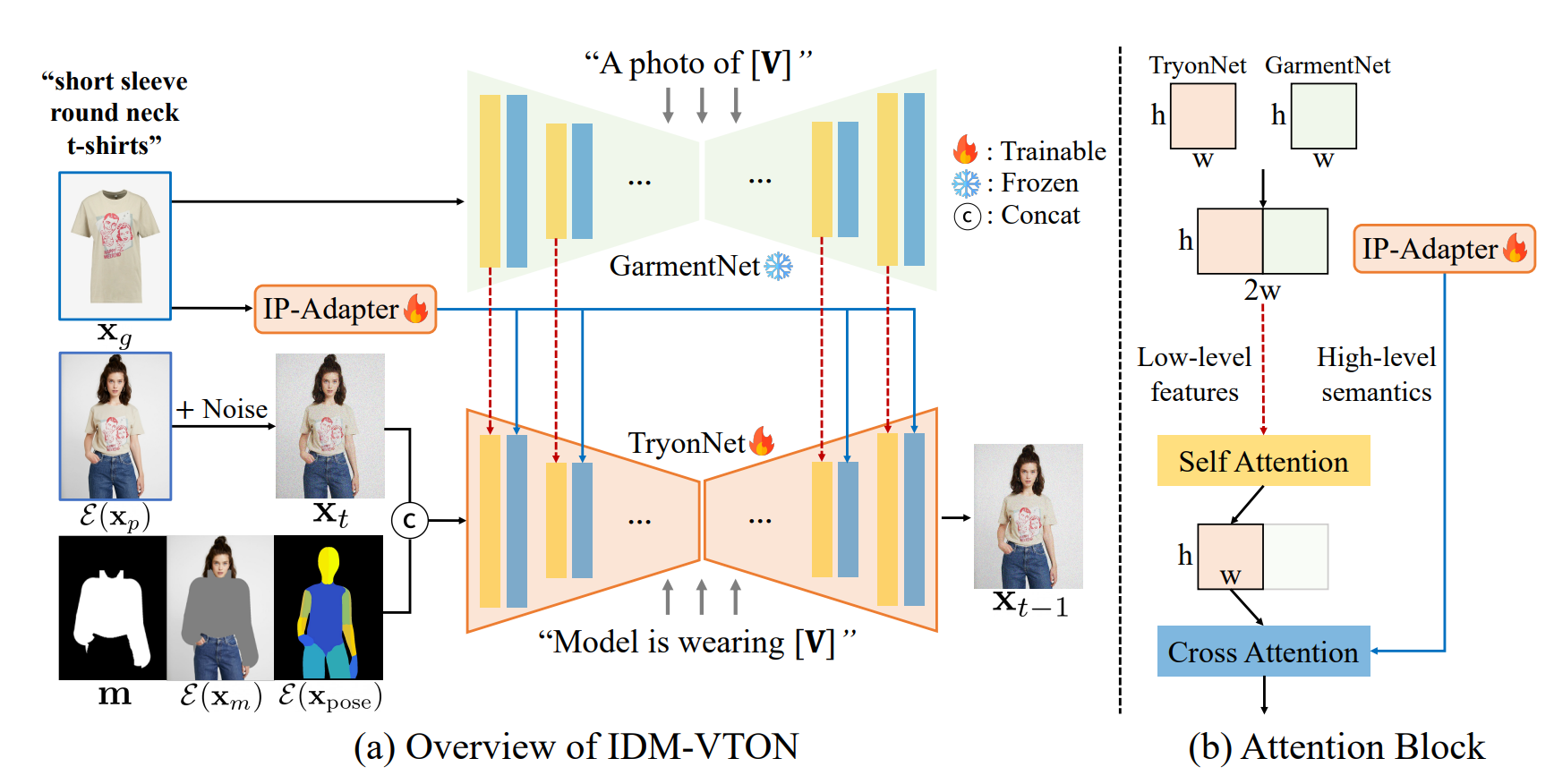}
    \caption{The architecture of IDM. Similarly to OOTD, it uses parallel unets and concatenates self-attention maps before doing the self-attention.}
    \label{idm}
\end{figure}

The main difference between the two is firstly the prevalence of an ip-adapter at IDM viton, but also the fact that the non-denoising unet is frozen. Having a frozen pre-trained image-to-image unet appears to be a good direction. Intuitively, the frozen unet maintains details about the garment whereas the trainable learns which of these details to transfer and whereto on the generated image.

\begin{figure}[H]
    \centering
    \includegraphics[width=0.8\textwidth]{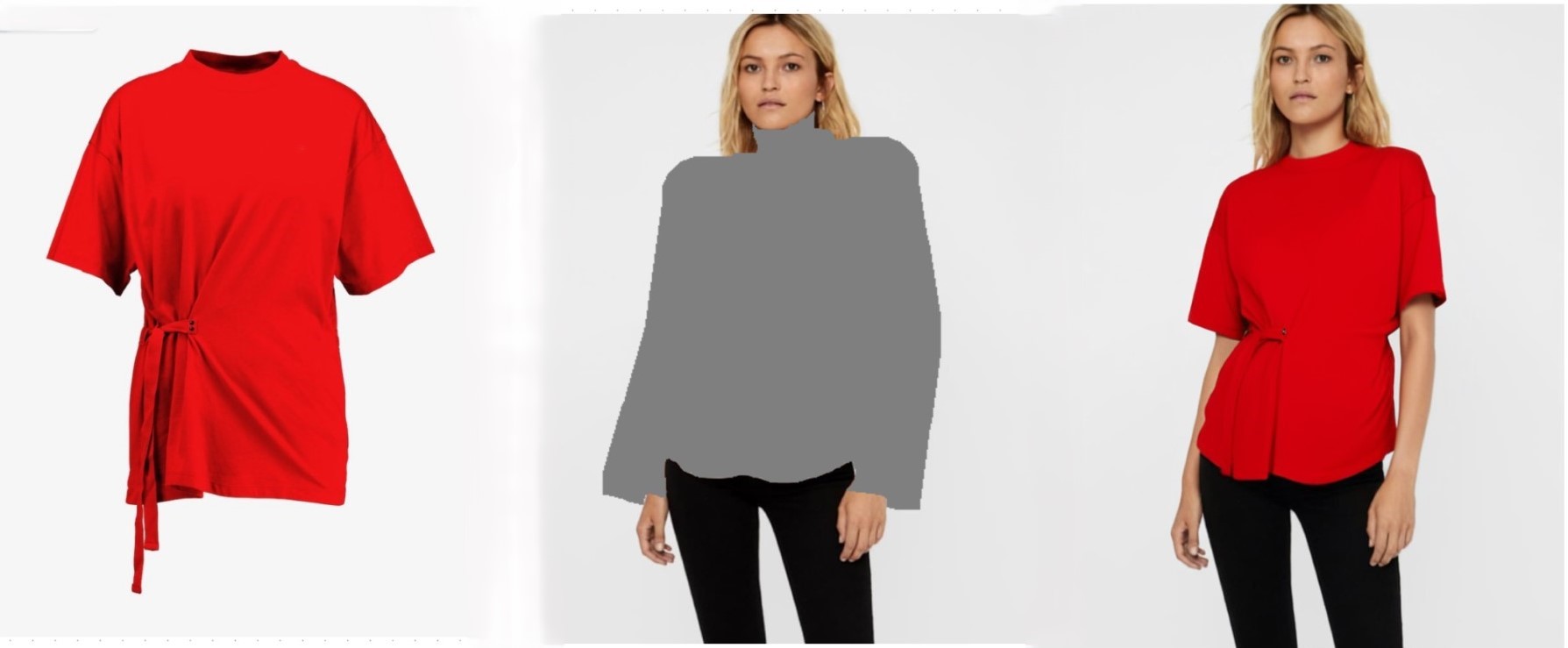}
    \caption{Example inference of IDM-viton, where the garment is mapped onto the masked area of the person. Note that it preserves details but also generates areas on arms such as the hem on arms and the forearms in their entirety.}
    \label{idm-test}
\end{figure}

In short, almost 2 years after stable diffusion's open release, many very useful and powerful models have been released that leverages the inherent generative capability but that also are finetuned for very specific purposes.

\section{Method}

The following section presents the background and leads into outlining the proposed method.

\subsection{Background on Diffusion Models}
Diffusion models are based on the principle of gradually adding noise to data and
then learning to reverse this process \cite{ho2020denoisingdiffusionprobabilisticmodels}.
We define $t=0$ as the image at the start of the process, i.e.\ unaltered, and $t=1$
as the image being pure noise. Letting $\mathbf{x}$ denote the image (a tensor), the
transformation that describes the diffusion process is defined as
\begin{equation}
    q(\mathbf{x}_t \mid \mathbf{x}_0)
    = \mathcal{N}\!\left(\mathbf{x}_t;\, \sqrt{\bar{\alpha}_t}\,\mathbf{x}_0,\,
      \left(1 - \bar{\alpha}_t\right)\mathbf{I}\right),
    \label{eq:forward_marginal}
\end{equation}
where $\bar{\alpha}_t \in [0,1]$ is a monotonically decreasing noise shedule with
$\bar{\alpha}_0 = 1$ and $\bar{\alpha}_1 = 0$. Equivalently, through the
reparameterisation trick, a noisy sample at time $t$ can be written as
\begin{equation}
    \mathbf{x}_t = \sqrt{\bar{\alpha}_t}\,\mathbf{x}_0
    + \sqrt{1 - \bar{\alpha}_t}\,\boldsymbol{\epsilon},
    \qquad \boldsymbol{\epsilon} \sim \mathcal{N}(\mathbf{0}, \mathbf{I}).
    \label{eq:reparam}
\end{equation}

The learning target of a diffusion model is, given a noisy image $\mathbf{x}_t$ and
the corresponding timestep $t$, to predict the noise $\boldsymbol{\epsilon}$ that,
according to Eq.~\eqref{eq:reparam}, most probably produced $\mathbf{x}_t$. Given a
model $\boldsymbol{\epsilon}_\theta$ parameterised by $\theta$ and a conditioning
signal $\mathbf{c}$, the training objective is the denoising score-matching loss
\begin{equation}
    \mathcal{L}(\theta)
    = \mathbb{E}_{\mathbf{x}_0,\, \mathbf{c},\, \boldsymbol{\epsilon} \sim
      \mathcal{N}(\mathbf{0}, \mathbf{I}),\, t}
      \left[\,
      \bigl\lVert \boldsymbol{\epsilon}
      - \boldsymbol{\epsilon}_\theta(\mathbf{x}_t, t, \mathbf{c}) \bigr\rVert_2^2
      \,\right],
    \label{eq:diffusion_loss}
\end{equation}
where $\mathbf{x}_t$ is constructed from $\mathbf{x}_0$ and $\boldsymbol{\epsilon}$
as in Eq.~\eqref{eq:reparam}.

By scaling and subtracting the predicted noise from $\mathbf{x}_t$, we retrieve the
model's estimate of the clean image,
\begin{equation}
    \hat{\mathbf{x}}_0
    = \frac{\mathbf{x}_t - \sqrt{1 - \bar{\alpha}_t}\;
      \boldsymbol{\epsilon}_\theta(\mathbf{x}_t, t, \mathbf{c})}
      {\sqrt{\bar{\alpha}_t}}.
    \label{eq:x0_estimate}
\end{equation}
Notably, since the scaling is continuous in $t$, we are not restricted to the fully
denoised endpoint: re-applying Eq.~\eqref{eq:reparam} to $\hat{\mathbf{x}}_0$ with
the same predicted noise yields an estimate at any intermediate noise level
$s \in [0, t]$,
\begin{equation}
    \hat{\mathbf{x}}_s
    = \sqrt{\bar{\alpha}_s}\,\hat{\mathbf{x}}_0
    + \sqrt{1 - \bar{\alpha}_s}\;
      \boldsymbol{\epsilon}_\theta(\mathbf{x}_t, t, \mathbf{c}),
    \label{eq:intermediate_estimate}
\end{equation}
which recovers $\hat{\mathbf{x}}_0$ for $s = 0$ and reproduces $\mathbf{x}_t$ for
$s = t$.

Most modern models, including Stable Diffusion, operate in a latent space rather than
directly on the pixels the user sees. This is considerably more efficient, as no
generative capacity is spent modelling the strong correlations between neighbouring
pixels~\cite{latent2022}. The model therefore learns to generate a latent
representation, which is subsequently mapped to pixel space by a variational
autoencoder (VAE).

A VAE consists of an encoder--decoder architecture that learns to compress
high-dimensional data into a lower-dimensional representation, which is in turn easier
to model generatively than a full-resolution image. The encoder, denoted
$\mathcal{E}$, maps an image $\mathbf{x} \in \mathbb{R}^{H \times W \times 3}$ to a
latent $\mathbf{z} = \mathcal{E}(\mathbf{x}) \in \mathbb{R}^{h \times w \times c}$ with
$h = H/f$ and $w = W/f$ for a downsampling factor $f$, while the decoder $\mathcal{D}$
performs the inverse mapping, such that $\mathcal{D}(\mathcal{E}(\mathbf{x})) \approx
\mathbf{x}$.

\subsection{Plug-ins for tuning diffusion-models}

Around foundational diffusion models, there exists an eco-system around them that enables specialized use cases, which can be described as plug-ins.

\subsubsection{Image prompt (IP) adapter }
The ip-adapter to condition the diffusion model on an image as opposed to only a text prompt. For both text an images, the CLIP text and image encoder is used \cite{radford2021learningtransferablevisualmodels}. 

\begin{figure}[H]
    \centering
    \includegraphics[width=0.6\textwidth]{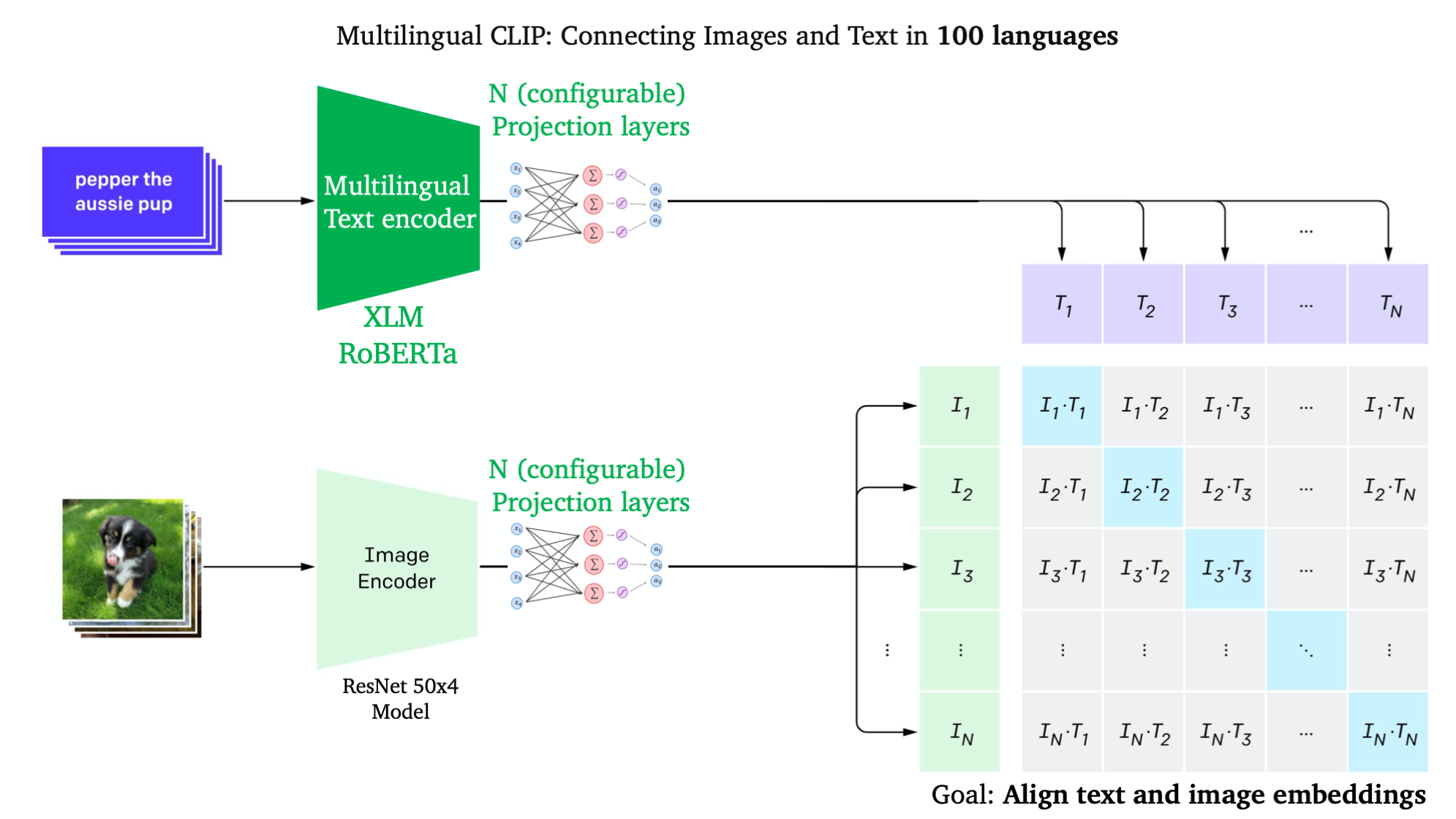}
    \caption{The CLIP text embedding that SDXL is trained with a corresponding CLIP image-embedding}
    \label{clip}
\end{figure}

Since the text and image prompt are contrasted to each other (maximizes dot-product), the embedding corresponding to an prompt of a description of a garment will by default be very similar to the image embedding of that garment. However, the image embedding will be able to capture more detail about the garment since it is processed from the image itself directly. The ip-adapter is trained on the output of the image encoder and inputs that for cross attention in the unet. 

Let \( Q \in \mathbb{R}^{N \times d} \) represent the query matrices from the intermediate
representation of UNet, and let \( K_c \in \mathbb{R}^{N \times d} \) and \( V_c \in \mathbb{R}^{N \times d} \) denote the key and value matrices derived from the text embeddings \( c \), where \( N \) is the number of samples. The output
of the cross-attention layer is defined as:

\[
\text{Attention}(Q, K_c, V_c) = \text{softmax}\left(\frac{Q K_c^\top}{\sqrt{d}}\right) \cdot V_c.
\]

Subsequently, the IP-Adapter computes the key and value matrices \( K_i \in \mathbb{R}^{N \times d} \) and \( V_i \in \mathbb{R}^{N \times d} \) from the image embeddings \( i \), and incorporates the cross-attention by elements-wise addition to the embedding from the attention from the text-prompt.
\\

\subsubsection{Text to Image (T2I) adapter}
The T2I adapter \cite{mou2023t2iadapterlearningadaptersdig}, is a light-weight trainable module that can be added to a denoising unet in order to condition it on an image. Differentiating it from the IP-adapter is that:
\begin{itemize}
    \item It is larger, around 100 million (depending on variant) compared to the IP which has around 22 million.
    \item The IP-adapter injects a tensor for cross attention layers, whereas the T2I adapter does element-wise addition to down sampling blocks in UNET
    \item The T2I-adapter's "bottle-neck" is in the order of several millions, meaning that at no layer is information compressed down very much. The IP adapter compresses the image into one or several 1280-dim embedding, meaning that there is a limit to how much detail it can capture for an image containing millions of pixels.
    \item The T2I adapter injects different values through-out the depth of the unet, where as for the IP-adapter, the same embedding is used for every cross-attention-layer

\end{itemize}

Combining these two into one may have several benefits:

\begin{itemize}
    \item The adapters can capture different aspects of the image. The IP-adapter more semantic information about the garment whereas the the T2I can provide more details
    \item Whereas the T2I adapter only can capture images, the fact that the embedding from the IP-adapter (can be) only on dimension, makes it easy to fine-tune its output through a MLP as well ass adding one-hot encoded class labels or other information. Since its a bottleneck of 1280, it is relatively easy to make sure that the added information to that embedding have an actual impact.
    \item The fact that there already exists pre trained IP-adapters for recreating images make them easy to fine-tune. Also, they make for a great point of initialisation for the model that helps the untrained T2I adapter lock on to a path of extracting detail.
\end{itemize}

Similar to many other methods for control, such as the control net \cite{control2023t2iadapterlearningadaptersdig}, the T2I-adapter focus on guiding semantic information, such as style or position of certain things. Whereas semantics are useful to some extent, we are just as interested in detail. This would mean having the adapter working on the up-blocks of the unet which corresponds to the final output, and the details in the image.

\subsection{Pre-processing}

The proposed method aims to make model able to do precise adjustments of garments. That which enables it is to have such precises dimensions as labels in training data. In short, we augment the input image of the garment with information about how should be aligned. 

\subsubsection{Parsing and segmentation}

As part of pre-processing of dataset, an important step is to create mask the images before feeding them into the model as seen in Figure \ref{idm-test}. With bad masking, the model with not learn to generate satisfactory images. Most previous models, including IDM and OOTD, have used Self-Correction-for-Human-Parsing \cite{parsing2022} for generating masking of images, since this was the original masking used for the the two big datasets available for academic use: DressCode \cite{dress2022dresscodehighresolutionmulticategory} and VitonHD-dataset\cite{choi2021vitonhdhighresolutionvirtualtryon}. Self-Correction-for-Human-Parsing is a model that both segments and classifies clothes and body parts on humans. Recently however, there have are newer models that put together can do the job in a more flexible way and also being completely open-source, paving the way for creating a VITON model for commercial use for any actor with a dataset.

One such model is GroundingDINO \cite{dino2024groundingdinomarryingdino}, released in 2023. For the purposes of this project, its a supercharged version of CLIP, that not only allows one to produce zero-shot classification of items, but also do so with regards to to positions in an image. 

Whereas GroundingDINO can label objects, it cannot segment them. Segmentation map is needed in order to pre-process the dataset. For this purpose, Segment Anything by Meta \cite{sam2023segment} can be used as an image segmentation model. These two models paired together allows for segmenting and label clothes by type, and found their exact bounding boxes which is central to the proposed method.

\subsubsection{Measuring distances on body}
The idea for this model is to utilize really good prepossessing in order improve results. Also, with the goal being able to control the outfitting of the garments, it is crucial that the model is trained om images that are labeled and processed in such a way that it "learns" the implications of these features.

One such feature is the garments position on the body. It was attempted to use chatGPT-4o vision for this purpose. However, models such as those are relatively bad at pinpointing exact positions of an item, such as the length of sleeves on an image since they operate on the latent space. For this method, new methods was used that leveraged the segmentation in order to give quantitative answers. The things measured were:

\begin{itemize}
    \item Garment length around waist (left and right side)
    \item Portion of shoulders that was bare
    \item Neckline depth
    \item Neckline Width
    \item Length of sleeves
\end{itemize}

Garment length around waist was decided using a combination of the segmentation with detectron's openspose model \cite{detect2023detectioninstancesegmentationclassification}. It functions by firstly identifying the hips on a person. If this point was located on the garment, the point "made a walk" towards the knee. When it was no longer on upper garment, it measured distance. On the other hand, if the point was not on the garment, it walk upwards until it was on the garment and measured the distance to hip key-point, as negative. This was both for left and right side. The distance was then normalized. First by dividing the measured length, (positive or negative), and dividing by the distance between chest key-point and hip key-point. This to ensure that varying sizes of people or how close the person is to the camera wont affect the measurement. The data was then normalized across the dataset.

\begin{figure}[H]
    \centering
    \includegraphics[width=0.85\textwidth]{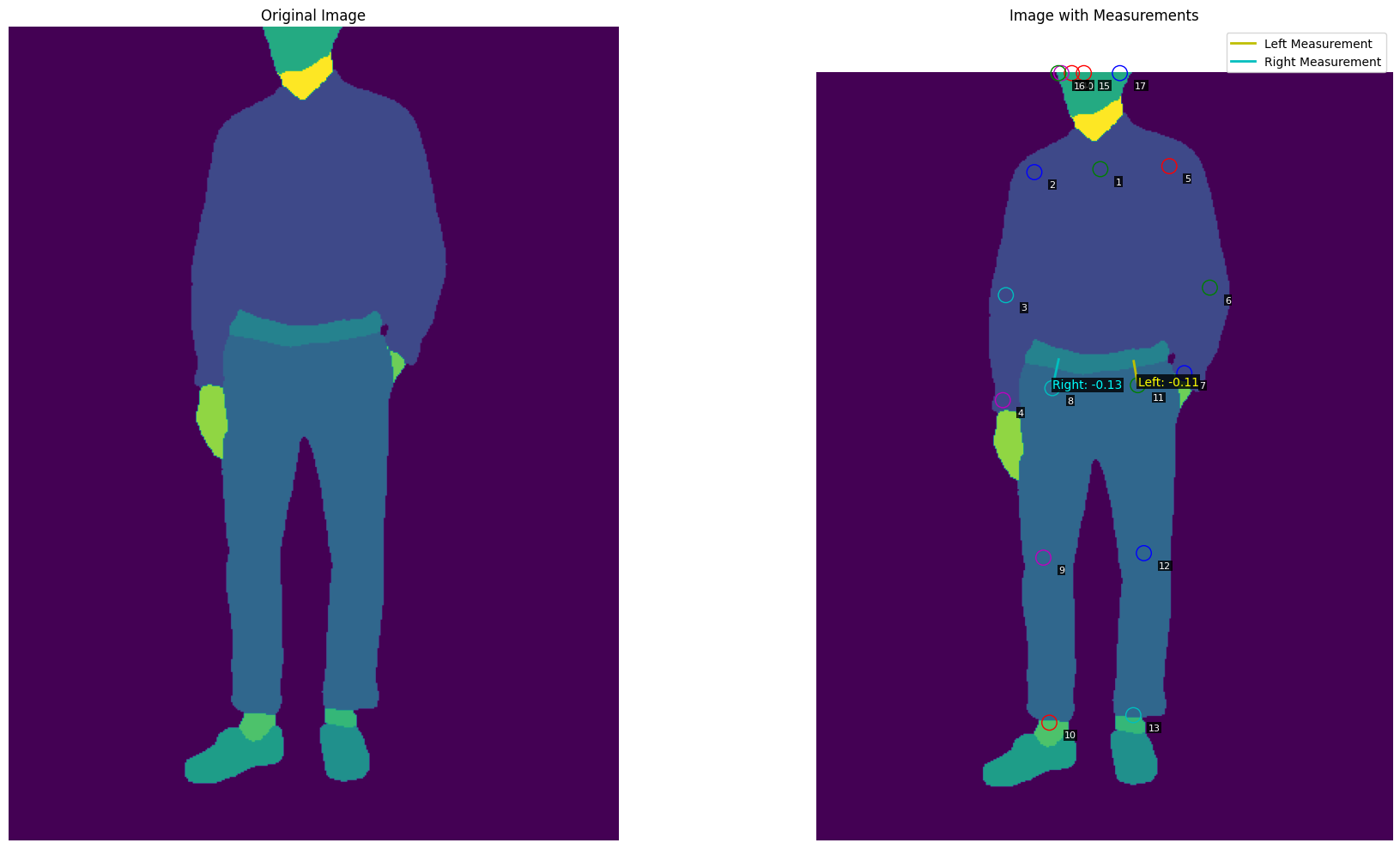}
    \caption{Visualisation of garment length}
    \label{gan}
\end{figure}

The amount of bare shoulders was decided by measuring if there was segmented garment on the shoulders. At the key-point, a normalized radius was taken and the fraction of pixels that was on the garment was measured. 

\begin{equation*}
    \text{fraction}=\frac{\text{garment area}}{\text{total area}-\text{background area}}
\end{equation*}
\\
The data was then represented as a float between 0 and 1 where 0 are bare shoulders and 1 are fully covered. The rationale for this approach is to capture situations where parts of shoulders are covered, as well as to make it more robust for errors in segmentation.

Neckline depth and width was done by measuring the distance if neckline in relation to the chest key-point. Firstly, all connected regions of segmented skin tha was not the chest and throat area was removed to make sure we were measuring the correct area. The y-coordinate of the chest key-point was chosen and from that, lowest point of visible skin indicated where neckline ended. If the person wore a turtle neck the measured distance would be between chest and chin whereas an open shirt would lead to a negative distance below the chest. The width mas measured by taking the x-distance between left and right most visible part of skin. The distances was normalized by upper body height and shoulder width respectively, and then normalized across the dataset.

\begin{figure}[H]
    \centering
    \includegraphics[width=0.5\textwidth]{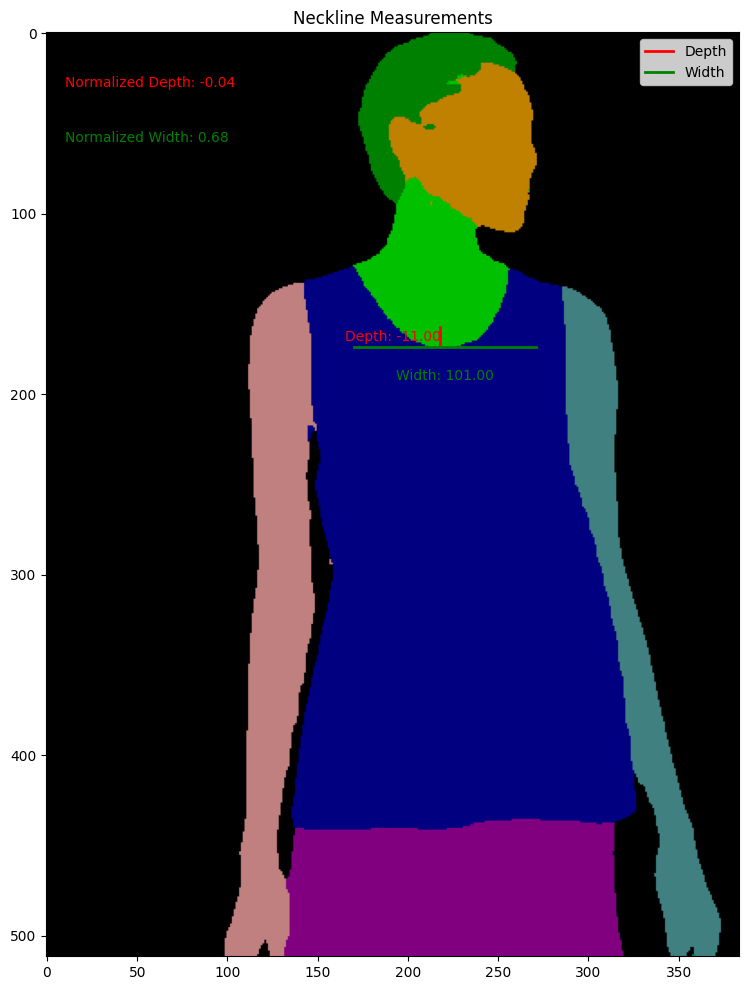}
    \caption{The green and red lines measure depth and widht of neckline}
    \label{gan}
\end{figure}

Length of sleeves was not measured as a distance, but rather as one-hot encoded classes with 6 varying lengths. It was done by:
\begin{itemize}
    \item Identify arm with most skin showed by area, some arms may be hidden from view and we only need to measure one.
    \item Identify the 3 key-points corresponding to shoulder, elbow and wrist
    \item Create two new intermediate points by measuring distance between these 3.
    \item Measure the lowest point at which there is garment, decided by taking an area around the point and seeing what fraction is garment, which is very robust.
    \item Assign class label for which sleeve length image belongs to.
\end{itemize}

\begin{figure}[H]
    \centering
    \includegraphics[width=0.5\textwidth]{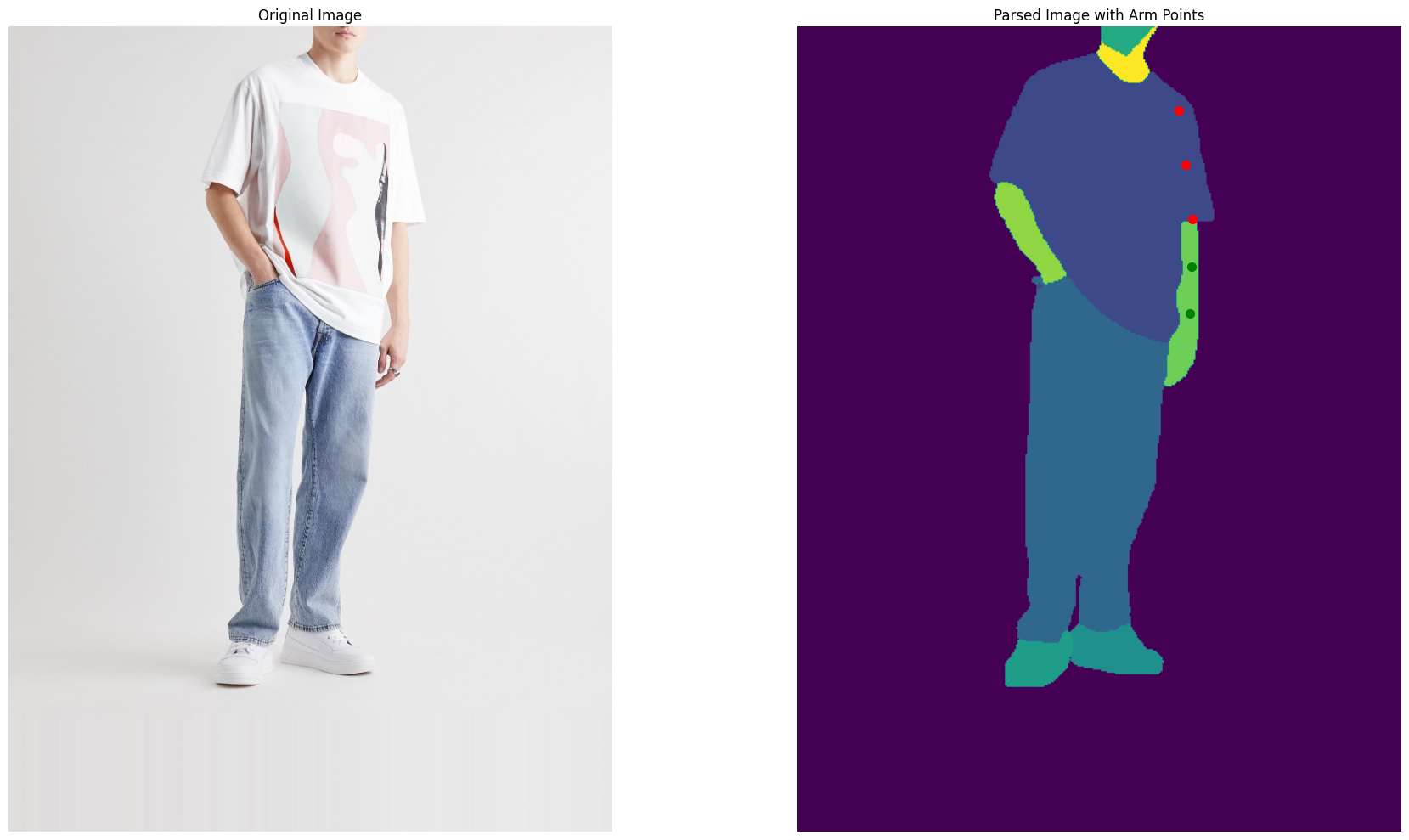}
    \caption{Red indicates that garment was detected}
    \label{gan}
\end{figure}

It is possible that it would be possible that using the previous techniques, i.e continuous measurement would have worked. But having two different method of measuring length allows for more evaluation. 
A big advantage is with continuous techniques is that it is much more dense as opposed to one-hot encoded classes. Also, at inference it is in theory possible to have more granular control. It is easy to change the number of neckline by sliding a number between values. Also, it potentially allows for "forcing" the model to behave in a certain way. If the model is not behaving as wanted, it is possible to set the parameter neckline-depth to -5 which corresponds to a depth of what would equate to 5 standard deviations lower neckline. This is outside the values of any data in the dataset bu may force the model to pay attention to that value. 

\subsubsection{Masking}

A crucial component of pre-processing the dataset is masking. Masking is aimed at removing the are to be in-painted. Whereas all previous models were trained using the same masked image for every item, the method used for COntrolableCLothing involves creating dynamic masks in the dataloader itself. Instead of pulling a masked image, the data-loader pulls the data it needs to create masked image. The mask is created can be therefor be slightly different every time, aiding the model to not over-fit to the hyper parameters of the masking the process, such as how fat down the mask goes down. The masing process works by:

\begin{itemize}
    \item Identifying the upper body from parsed image.
    \item Masking the garment and forearms
    \item Masking masking an area around arms masking a fix distance from the key-points of the arms. This mask ignores garment and "hides" information from the model
    \item Mask throat and area around head
    \item Add random number of pixels for mask all around for randomness
\end{itemize}

\begin{figure}[H]
    \centering
    \includegraphics[width=0.85\textwidth]{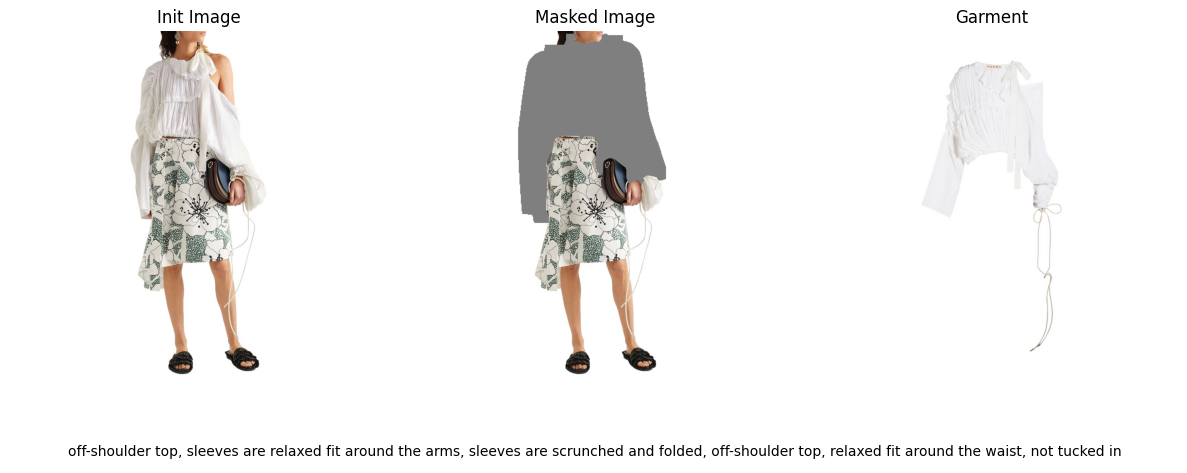}
    \caption{ Example of masked Image }
    \label{masked}
\end{figure}

What the masking operation also does is to allow for adding extra mask below the waist. If the mask ends precisely at garment model learns to follow line of mask, which may or may not be desirable. For this reason the model is also given information about mask at training. It is given a boolean (0 or 1) if extra mask was added. It can then learn to generate extra distance from the mask. At inference, this could potentially give control over in-painting in relation to mask. (Figure \ref{masked} show no extra mask for waist.)

\subsection{Labeling dataset}
The final pre-processing step involves labeling the dataset. By default, the data is not labeled in any way. This means that models trained only on the images, can learn to transfer detail but the user can not specify the fit and other aspects of the garment. Even though Stable Diffusion (the base for almost all recent VITON) is multi-modal by default, the fact that no useful prompt was given during training results in the model ignoring the cross attention that attends to text embedding. The result is an "average" fit for all garment.

\begin{figure}[htb]
\centering
\begin{minipage}{0.44\textwidth}
    \centering
    \includegraphics[width=\textwidth]{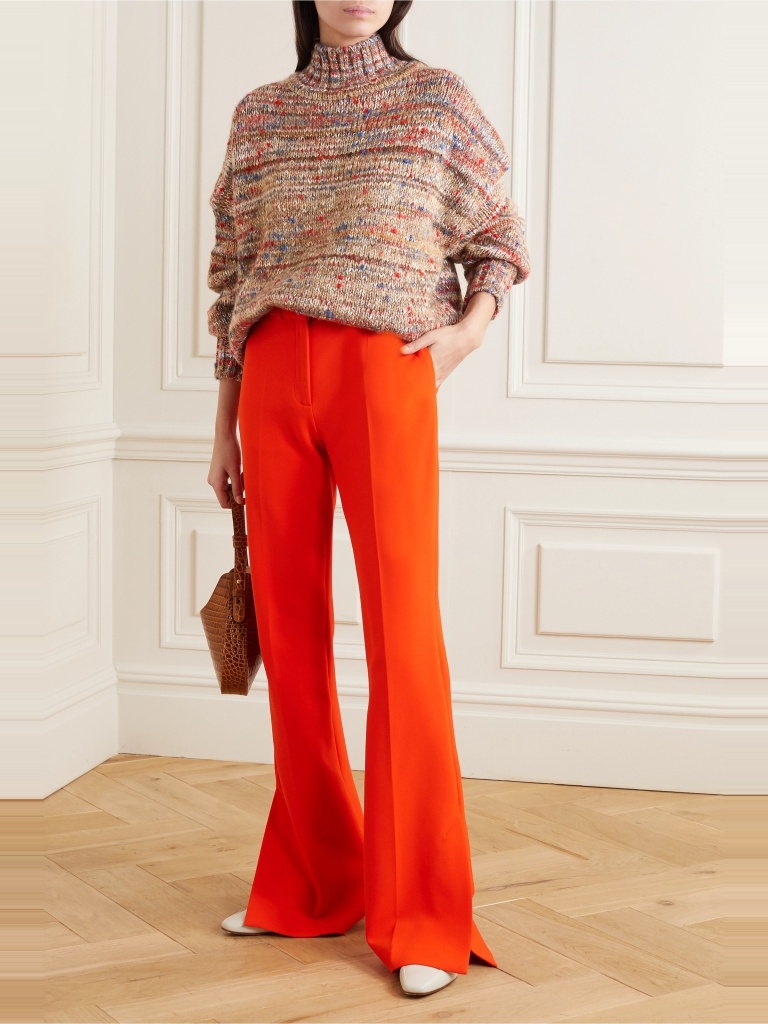}
    \caption{Reference image in dataset}
    \label{fig:your_image}
\end{minipage}\hfill
\begin{minipage}{0.54\textwidth}
    \centering
    \small
    \begin{tabular}{|p{0.65\textwidth}|p{0.3\textwidth}|}
    \hline
    \textbf{Question} & \textbf{Answer} \\
    \hline
    How is the upper garment fitted around arms? & Relaxed \\
    \hline
    How is the upper garment fitted around chest/shoulders for its type? & Relaxed \\
    \hline
    How is the upper garment fitted around waistline? & Relaxed \\
    \hline
    Where does sleeve-hem or cuffs extend to on arms, closest points? & Elbow \\
    \hline
    Can the upper garment be considered 'over-sized'? & Yes \\
    \hline
    How snugly does sleeve-hem fit? & Loose \\
    \hline
    How is tucking of upper garment? & Loose tuck with blousing \\
    \hline
    Does garment cover entirety of shoulders? & Both \\
    \hline
    What describes styling of cuffs? & Elastic or Ribbed cuffs \\
    \hline
    What describes sleeves best? & Full-length \\
    \hline
    Are sleeves slightly scrunched or pushed back on forearm? & Yes \\
    \hline
    What describes the neckline type the best? & Turtleneck \\
    \hline
    Collar type? & Neckline/Other Descriptor \\
    \hline
    Type of upper garment? & Pullover \\
    \hline
    How far down does the upper garment extend? & To the hips \\
    \hline
    How are the buttons fastened in front? & Does not have buttons \\
    \hline
    \end{tabular}
    \captionof{table}{Questions and inferred category using gpt-4o}
    \label{tab:image_analysis}
\end{minipage}
\end{figure}

Recent advancements in large multi-modal models, such as GPT-4o, allows for accurate labeling of datasets, to an affordable costs. In Figure \ref{tab:image_analysis}, one cans an image paired with questions asked about it. The model was asked to select between a selection of classes for each question, that the could be turned into a one-hot encoded embedding. The labels could also be used for prompting by turning the several labels into a sentence of of fit of garment. 
The datasets used were Dresscode \cite{dress2022dresscodehighresolutionmulticategory} and VitonHD \cite{choi2021vitonhdhighresolutionvirtualtryon}, that both contain around 15k images of pairs between a garment and a person wearing it.

\section{Proposed method}

\begin{figure}[H]
    \centering
    \includegraphics[width=0.85\textwidth]{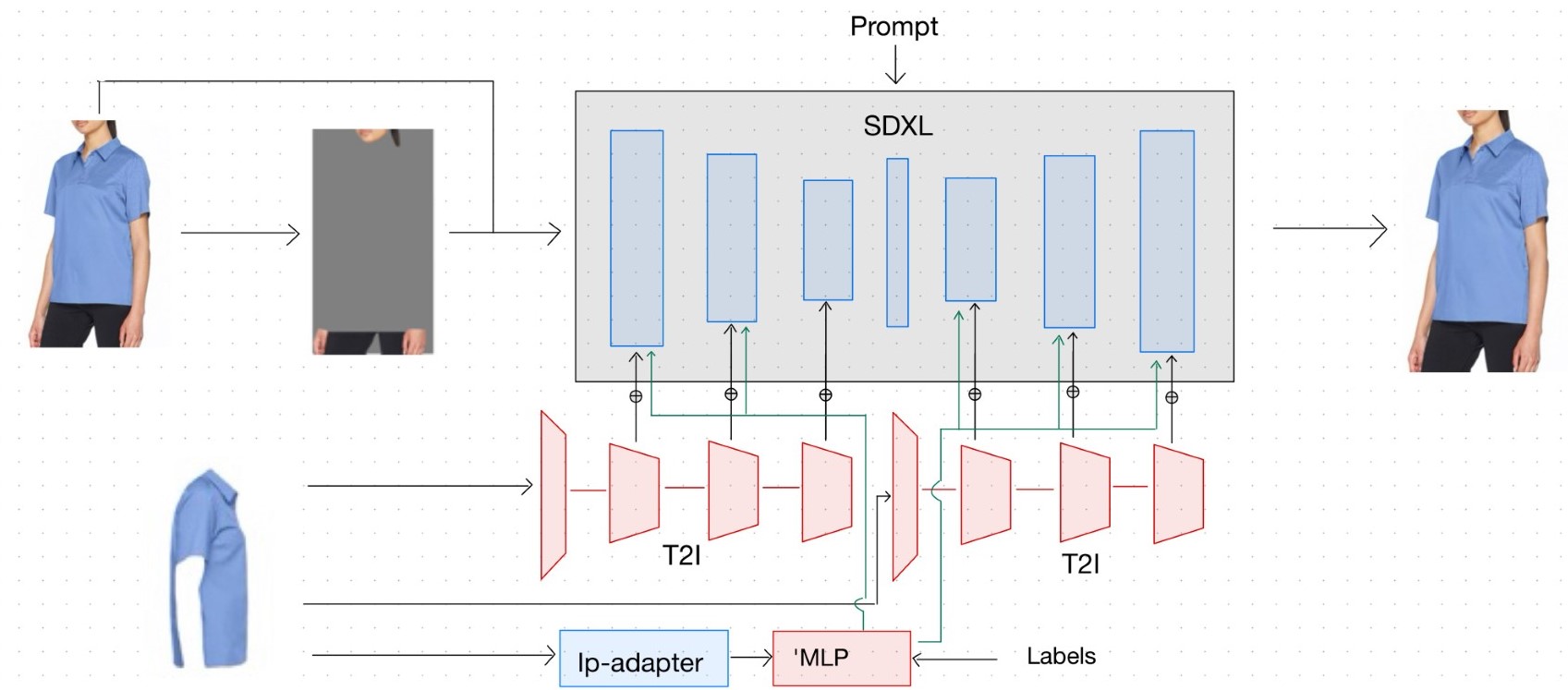}
    \caption{The architecture of controllable clothing. The trainable components are the T2I adapters and the fine-tuning MLP for the IP-adapter, that also incorporates the labels and information about the mask. }
    \label{gan}
\end{figure}

\subsection{Overall Structure}

The proposed methods for injecting the new information is by adding it tho the input embedding for the cross attention. Whereas the embedding holds semantic information about the garment, it lack any information about how it should be styled. The trainable module is in effect a resNet. As input it had the embedding as well as the tensor containing labels and distances of garment. It puts these trough a MLP that then is element-wise added to the original embedding.

For more detailed information about the garment, T2I adapters were used. The down-sapling T2I is an established module but there existed no similar module for up-sampling blocks, which had to be custom-created to fit the dimensions of the intermediate tensors between the blocks.

\subsection{Custom T2I for up-sampling}

For capturing detail in the garment, information must be injected into the up-blocks, where details are produce. The fact that a normal T2I adapter doe snot suffice was apparent after a few epochs, and the subsequent additional module was added at epoch 17.
Most methods use parallel unets which are computationally expensive. During both training and inference, both memory usage and computationally cost is doubles as an identical architecture runs in parallel for every time-step.  For this reason a lighter adapter was added with approximately 72 million parameter that does element-wise addition to the intermediate tensors between the blocks, having dimensions:
\begin{itemize}
    \item \([ \text{batch\_size}, 1280, \text{Resolution\_y}/16, \text{Resolution\_x}/16 ]\)
    \item \([ \text{batch\_size}, 640, \text{Resolution\_y}/8, \text{Resolution\_x}/8 ]\)
    \item \([ \text{batch\_size}, 320, \text{Resolution\_y}/8, \text{Resolution\_x}/8 ]\)
\end{itemize}

This is only done once before denosing loop and is very cheap. In theory, it does not have the same capabilities for detail, but for evaluating semantic features it may work sufficiently well.

It is also worth noting that no preexisting pipelines for ant T2I adapter existed in pair with in-painting and ip-adapter that I could find and had to be created from scratch. In the case for up-sampling T2I adapter, that module was created from scratch and added to the pipeline.

\section{Evaluation and results}

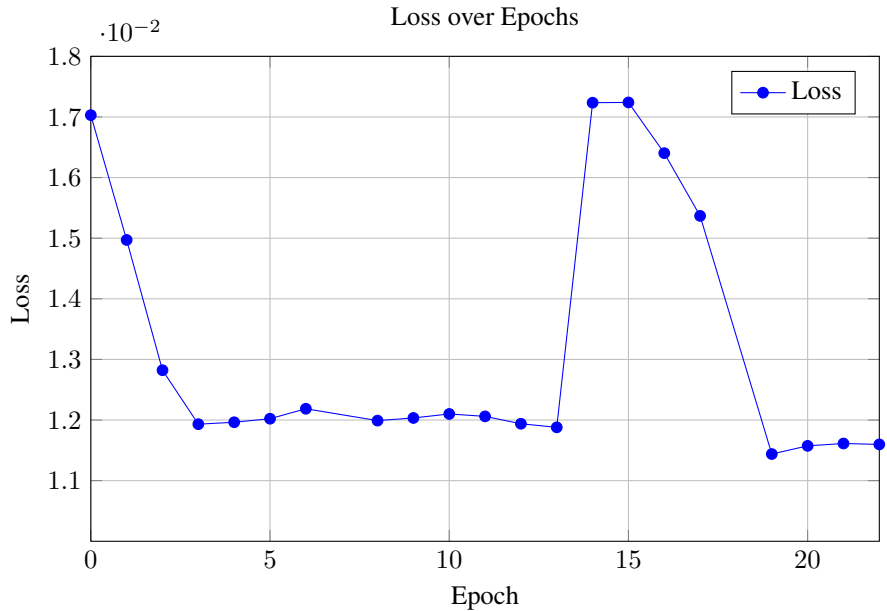
\begin{figure}[H]
    \centering
    \begin{tikzpicture}
\begin{axis}[
    width=12cm,
    height=8cm,
    xlabel={Epoch},
    ylabel={Loss},
    title={Loss over Epochs},
    xmin=0, xmax=22,
    ymin=0.01, ymax=0.018,
    xtick={0,5,10,15,20},
    ytick={0.011,0.012,0.013,0.014,0.015,0.016,0.017,0.018},
    yticklabel style={/pgf/number format/.cd,fixed,precision=3,/tikz/.cd},
    grid=major,
    legend pos=north east,
    legend entries={Loss},
]
\addplot[
    blue,
    mark=*,
    ]
    coordinates {
    (0,0.01703011194451257)
    (1,0.014971088377158984)
    (2,0.012821905668478317)
    (3,0.011931604274930936)
    (4,0.011963508620848497)
    (5,0.012021823602600692)
    (6,0.012185854964544808)
    (8,0.011990537873009388)
    (9,0.012034172350970972)
    (10,0.012100187461719003)
    (11,0.012060486812752878)
    (12,0.011938627462691118)
    (13,0.011879669665980044)
    (14,0.017235454035352576)
    (15,0.017239639468653825)
    (16,0.016402846651666918)
    (17,0.015367780672007692)
    (19,0.011439027467139698)
    (20,0.01157419907151726)
    (21,0.011612961844093056)
    (22,0.011597858223052938)
    };
\end{axis}
\end{tikzpicture}
    \caption{MSE-Loss over Epochs, note that the VITON dataset was added at epoch 17 which lead to the big increase in loss}
    \label{fig:loss-over-epochs}
\end{figure}

The training process took a few days on an A100, during the time of which I was able to add more data while the run was going on. We see that the loss has a floor of approximately 0.012 under which it cannot overfit. This is likely due to the relatively low parameter-count acting as a regularizer.

\subsection{Visual Examples}

\begin{figure}[H]
    \centering
    \includegraphics[width=0.85\textwidth]{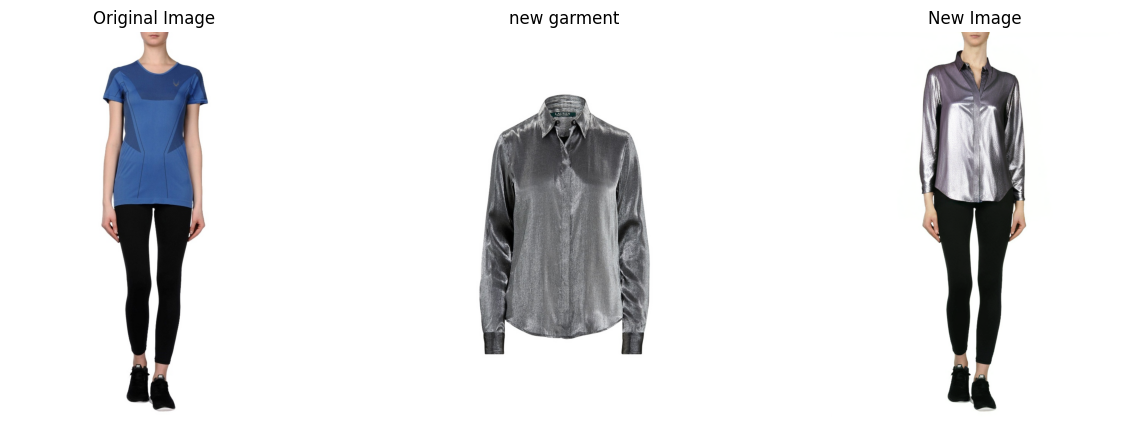}
    \caption{Example VITON using controllable clothing}
    \label{input}
\end{figure}

\begin{figure}[htb]
\centering
\begin{minipage}{0.38\textwidth}  
    \centering
    \includegraphics[width=\textwidth]{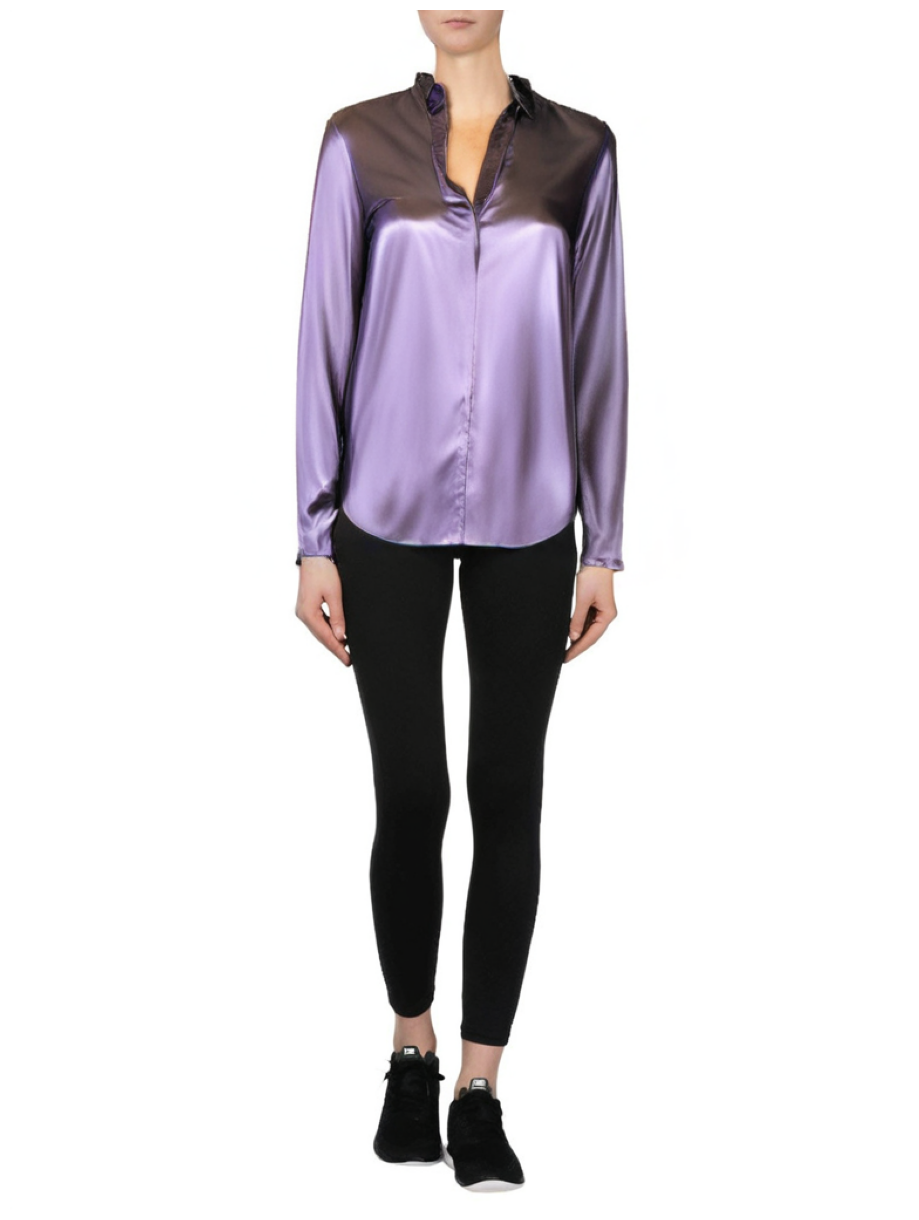} 
    \caption{Garment Image}
    \label{fig:purple}
\end{minipage}\hfill
\begin{minipage}{0.5\textwidth}  
    \centering
    \scriptsize  
    \begin{tabular}{|p{0.6\textwidth}|p{0.35\textwidth}|}  
    \hline
    \textbf{Condition} & \textbf{Value} \\
    \hline
    Sleeve Length & Elbow-length \\
    \hline
    Left Frac & 0.0 \\
    \hline
    Right Frac & 0.0 \\
    \hline
    Left Hip Dist & 0.0 \\
    \hline
    Right Hip Dist & 0.0 \\
    \hline
    Normalized Neckline Depth & 1 \\
    \hline
    Normalized Neckline Width & 1 \\
    \hline
    Arm Fit & Sleeves are relaxed fit around the arms \\
    \hline
    Waist Fit & Relaxed fit around the waist \\
    \hline
    Button Style & Several buttons are unfastened \\
    \hline
    Cuff Style & Rolled-up cuffs at the wrist \\
    \hline
    Is Oversized & It is oversized \\
    \hline
    Sleeve Scrunch & Sleeves are scrunched and folded \\
    \hline
    \end{tabular}
    \captionof{table}{Garment Conditions}
    \label{tab:garment_conditions}
\end{minipage}
\end{figure}

Figure \ref{input} show a relatively good example. Reason being that the garment is quite plain which make sit so that the semantic representation is enough to generate a believable image. The image generated has no prompt or class label and is using only the T2I adapters and ip-adapter. The following images show generated images using varying different settings for class labels but no prompt. As can be seen in Figure \ref{fig:purple}, manually changeling inputs may have unforeseen consequences like in this case where the garment turns purple.

\begin{figure}[htb]
\centering
\begin{minipage}{0.4\textwidth}  
    \centering
    \includegraphics[width=\textwidth]{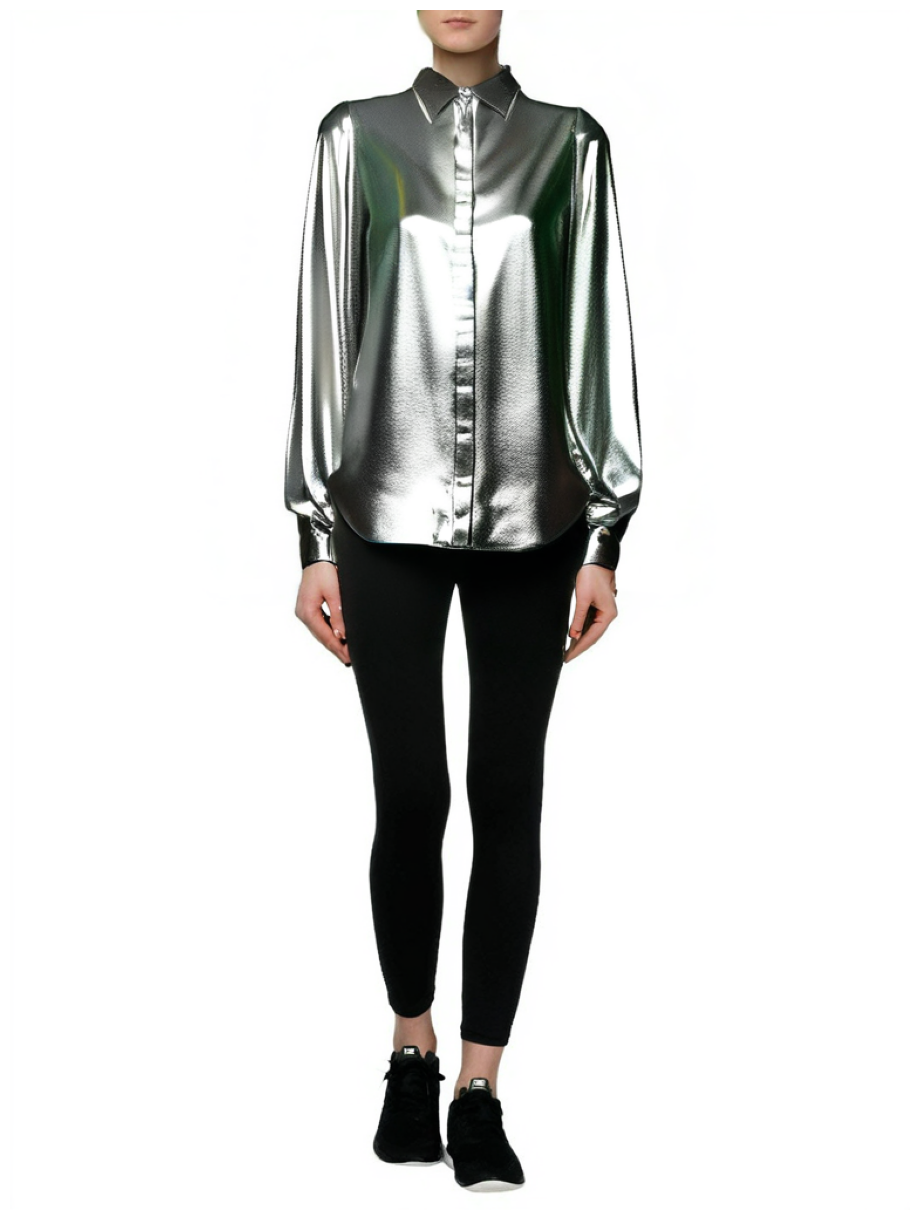} 
    \caption{Garment Image}
    \label{fig:tight}
\end{minipage}\hfill
\begin{minipage}{0.5\textwidth}  
    \centering
    \scriptsize  
    \begin{tabular}{|p{0.6\textwidth}|p{0.35\textwidth}|}  
    \hline
    \textbf{Condition} & \textbf{Value} \\
    \hline
    Sleeve Length & Full-length extended sleeves \\
    \hline
    Left Frac & 0.0 \\
    \hline
    Right Frac & 0.0 \\
    \hline
    Left Hip Dist & 10 \\
    \hline
    Right Hip Dist & 10 \\
    \hline
    Normalized Neckline Depth & -2 \\
    \hline
    Normalized Neckline Width & -2 \\
    \hline
    Arm Fit & Sleeves are tight fit around the arms \\
    \hline
    Chest/Shoulder Fit & Tight fit around the chest and shoulders \\
    \hline
    Waist Fit & Tight fit around the waist \\
    \hline
    Tuck Style & Fully tucked in \\
    \hline
    Button Style & All buttons are fastened \\
    \hline
    Cuff Style & Buttoned cuffs at the wrist \\
    \hline
    Is Oversized & Normal fit \\
    \hline
    Type & Plain sleeves \\
    \hline
    Sleeve Scrunch & Sleeves are scrunched and folded \\
    \hline
    \end{tabular}
    \captionof{table}{Garment Conditions}
    \label{tab:garment_conditions}
\end{minipage}
\end{figure}

As can bee seen in Figure \ref{fig:purple} and \ref{fig:tight}, changing the class labels can change the output the image the desired way. Empirically, the labels being numeric, i.e not one-hot encoded vectors, work better that one-encoded classes. Comparing the two images, we can see that both garment length and neckline was clanged from original in Figure \ref{input}. The controlling may however require numbers (standard deviations) much larger than that found in dataset. One possible explanation is that it is more accurate that one hot encoded image, but also, it always has a value. Making something that can be an number to one hot means that that attribute is sparse, which makes it more difficult for the nlp to learn it. In general with one-hot encoded vectors, there is always a trade-off between fidelity and sparsity. More classes leads to more fidelity but also more sparsity. \\
In the case of this model, it can also be prompted with text.

\begin{figure}[H]
    \centering
    \includegraphics[width=0.4\textwidth]{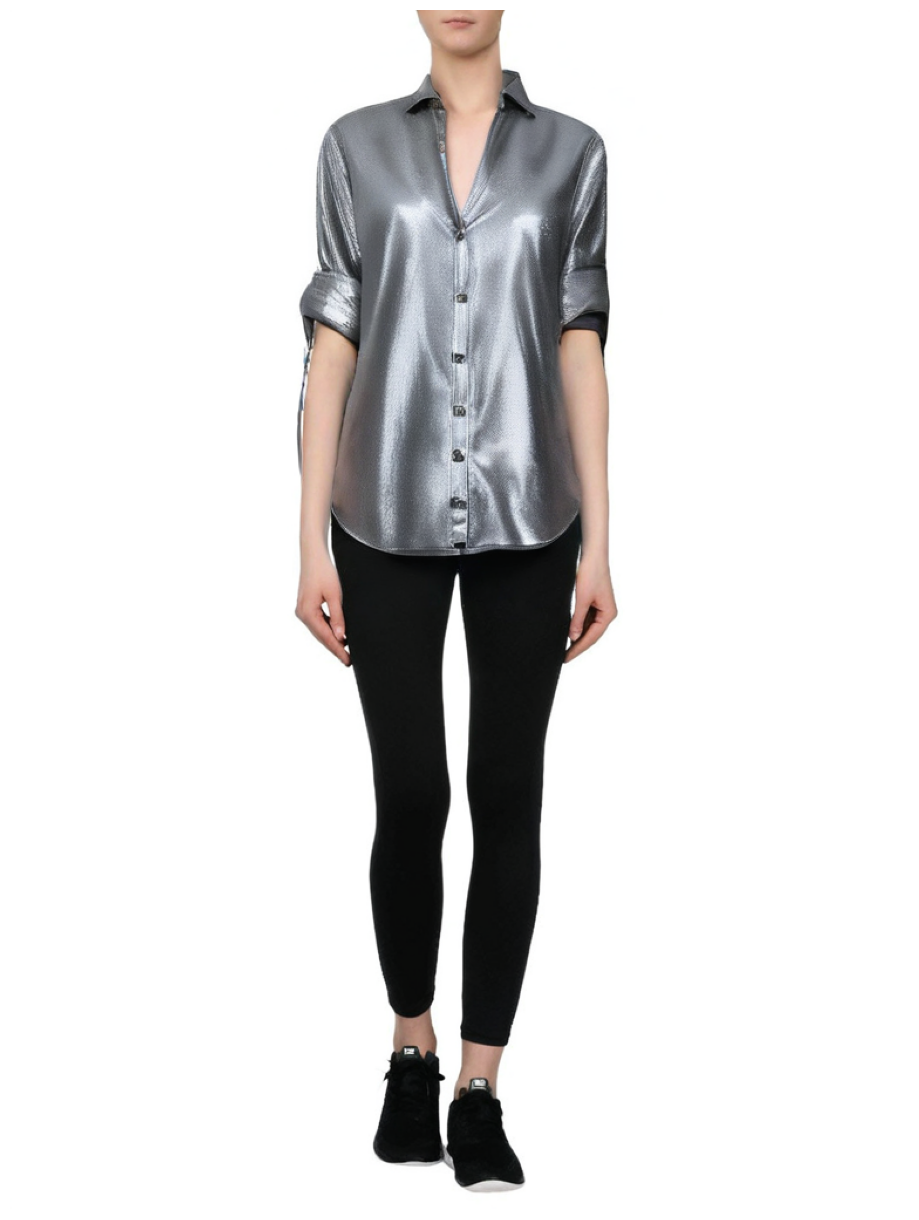}
    \caption{Same labels as Figure \ref{fig:purple}, but propted to be grey and rolled up sleeves.}
    \label{input}
\end{figure}

\subsection{Semantics and details}

With limited training, the model struggles with details. Whereas the model as whole has run 24 epochs, 7 of which on the whole dataset, the up-block T2I adapter was only added at epoch 17 as well. So it has also ran only 7 epochs, whereas hundreds is a more common number for these types of datasets.

\begin{figure}[h!]
    \centering
    \subfigure{}
    \includegraphics[width=0.49\textwidth]{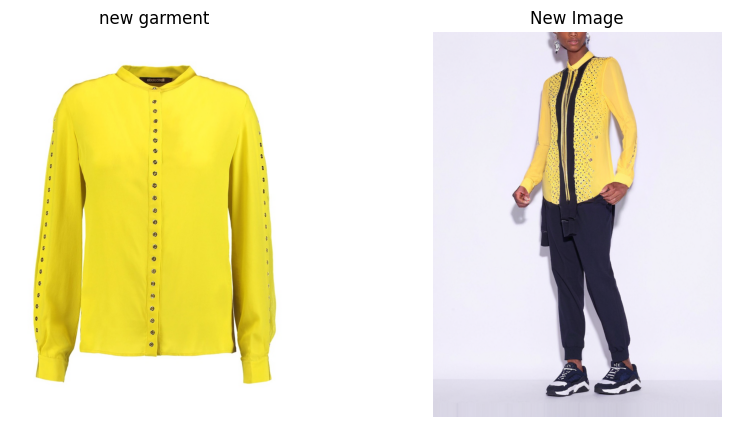}
    \subfigure{}
    \includegraphics[width=0.49\textwidth]{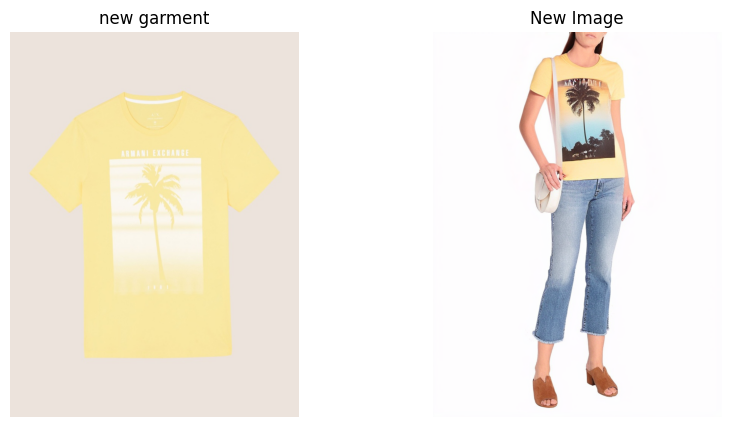}
    \caption{For highly detailed garments, detail may be lost unless there is a more trained module for up-blocks in place. interestingly, the image on the right have colours wrong but the right motive.}
    \label{bad}
\end{figure}

\begin{figure}[h!]
    \centering
    \subfigure{}
    \includegraphics[width=0.49\textwidth]{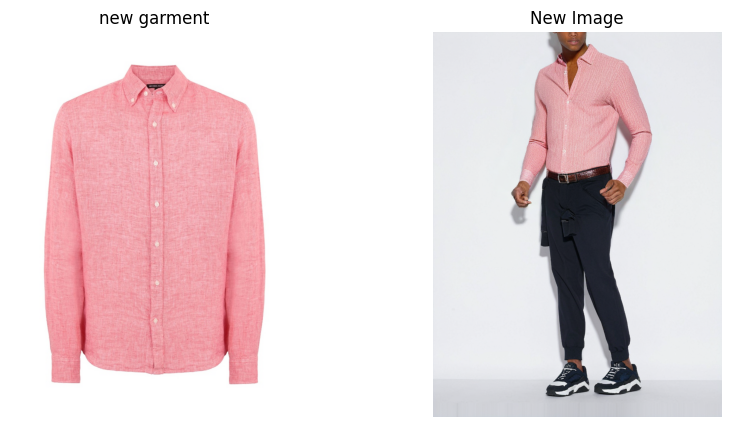}
    \subfigure{}
    \includegraphics[width=0.49\textwidth]{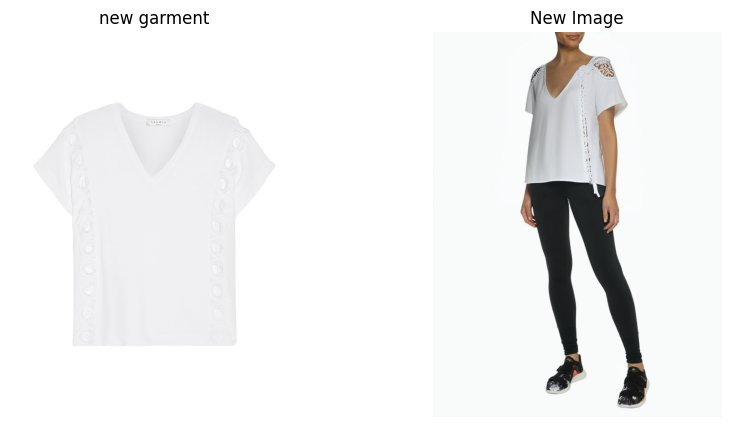}
    \caption{On the other hand, for more plain images, without much intricate detail, the model may produce images. The model need more training to get to a state where it reproduces detail.}
    \label{good}
\end{figure}

\newpage

\subsection{Shoulder fraction}

\begin{figure}[h!]
    \centering
    \subfigure{}
    \includegraphics[width=0.49\textwidth]{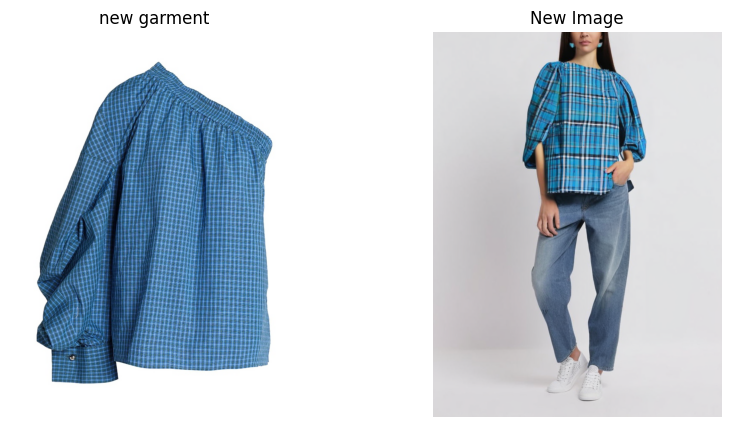}
    \subfigure{}
    \includegraphics[width=0.49\textwidth]{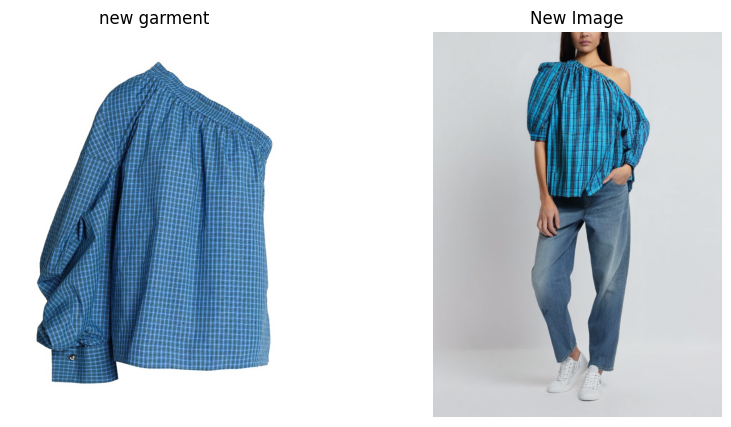}
    \caption{Setting different fractions for shoulders yields different styling of off-shoulder garments}
    \label{frac}
\end{figure}

Recall that the two parameters frac, left an right, corresponds to percentage of shoulder that is covered in garment. Observing Figure \ref{frac}, we note that the try-on by default does not have the garment as off-shoulder. By setting right-frac to 1 and left to 0, we tell the model to cover one shoulder but not the other. Note that no prompting was done for this image, only changing the two fraction-parameters.

\subsection{Control generation in relation to mask}

\begin{figure}[H]
    \centering
    \makebox[\textwidth][c]{%
        \includegraphics[width=1.3\textwidth]{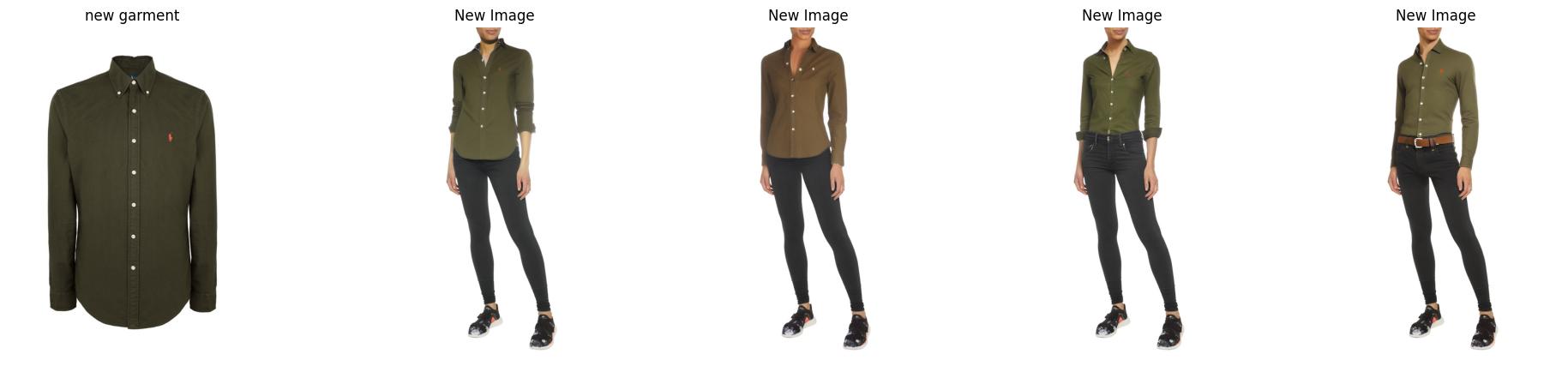}
    }
    \caption{Setting the parameter waist-mask to true or false yields different generation around waist.}
    \label{tuck}
\end{figure}

Another parameter for inference is waist-mask. In training, with a certain chance, extra mask is added to the image to the waist, forcing the model to regenerate the waist area. The model get the one-hot encode if this has happened or not. For a added mask, on can set that parameter to one or zero. These images had the exact same mask but the two on the left had waist-mask set to 0. This lead the model to create garment at the border of the mask. That's why the hem follows the same line. \\
On the other hand, the other two images on the right had waist mask set to 1. This led the model to regenerate waist area and produce two different results. \\
\\
This feature is potentially very useful. Lets say you have an image of a person wearing a low hanging shirt, and want to dress that person in a crop-top, you would have to mask the entire shirt. However, if the model is trained to only fill in mask starting from the bottom as in Figure \ref{idm}, that would be impossible. Using the method of ControllableClothing, one could be able to:
\begin{itemize}
    \item Select to regenerate from mask the area upwards.
    \item Set hip-dist to specified height to decide where upper garment should start.
    \item Generate image and have the area of waist and stomach unpainted by SDXL, leveraging its pre-trained generative capabilities.
\end{itemize}
This condition was only done for waist but can in theory be extended to other areas if desired.

\section{Ablation Study}

\begin{figure}[H]
    \centering
    \makebox[\textwidth][c]{%
        \includegraphics[width=1.3\textwidth]{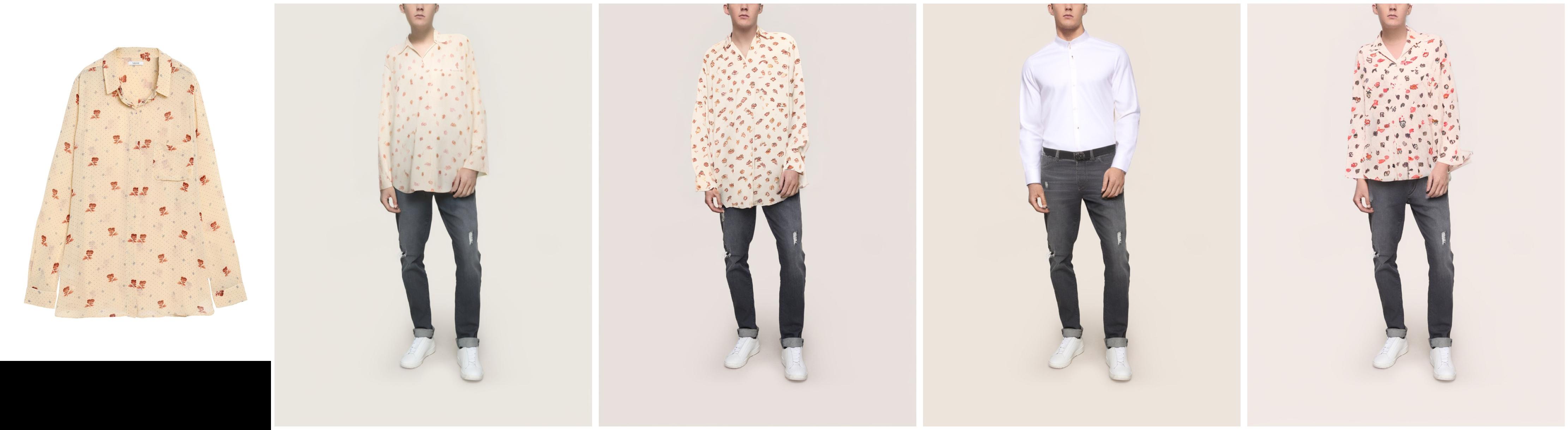}
    }
    \caption{From left to right, we have \\
    Original garment, generated image using all components, no up-block T2I, no IP-adapter and no down-block T2I}
    \label{abla}
\end{figure}

We note that all three images have an impact. Starting from the left, we note that the up-adapter arguably has the lowest impact. This may be due to it having a third the number of epochs trained with it. So drawing to many conclusions is premature. However it is interesting to note that compared to the image with all adapters, the one without up-T2I has a smaller density of "dots" compared to the others. This is detail that must be capture in the up-part of the unet so it would make sense for that adapter to have that effect. \\
The second image proves why the IP-adapter is such a crucial role. It initiates the generation to which the other modules can can add detail. It also explains why training was, all things considered, very efficient. Since the image was ok-looking from epoch 1, the other modules did not have to learn any semantics from garment, and can just focus on detail. So when there is no semantic to optimize it collapses.
Finally we have have the down-T2I. The apparent change is that the garment looses its colour  grading, and if one looks closely the collar changes from a normal collar to a revere collar. In short, at current state of training, it captures colours and some other semantic information.The colour grading makes sense as the IP-adapter may not encode a lot of colour on a limited vector.

\section{Discussion and future work}

In this study, the incorporation of extra pre-processing and labeling into a VITON model was examined. Empirically, using one-hot encoded features for semantic information is not ideal, since this information can be passed as text, which the model understands naturally.

At the same time, the effect of the one-hot encoding is unclear. During training, the type of garment (e.g., shirt, crop-top) was passed as a prompt along with five other random labels. It is plausible that the one-hot encoding did not contribute much, as the model already received the information as a string. This extends to sleeve length, where information such as "full-length extended sleeves" was added as a prompt. This may explain why the one-hot encoding did not improve performance despite conveying detailed information. In the future, this information should be encoded as a continuous variable between 0 and 1, corresponding to the sleeve length.

On the other hand, adding information that relates to precise boundaries is a good idea. Firstly, this is something that latent diffusion models struggle with. This was apparent when I attempted to have GPT label various positions of objects, and it performed poorly. Table \ref{tab:image_analysis} shows these questions being asked (such as sleeve length), but in the end, those types of questions were rejected. A custom method that utilized the segmentation map in pixel space, where distances are easier to measure, was adopted instead.

Remarkably, these features work quite well despite minimal training. A hypothesis for this is that they have a significant impact on the loss function. If the model generates an image of an upper garment where pants should have been, or some other misclassification occurs, it would significantly impact the loss, as opposed to small details on a garment.

The major benefits of this approach are that the model may potentially train faster since it is given explicit guidance on exact locations. It may also overfit less since fewer epochs are needed to achieve the same result. Additionally, it allows for more control during inference. This is especially true for measurements made in pixel space, such as relative distances, which latent models have difficulty with. Furthermore, many aspects such as the length of a garment and the depth of a neckline depend on styling and are impossible to predict solely from the mask and garment. For a new garment during training, it is challenging to guess how it is styled and fitted. Providing this information allows the model to focus on the relevant aspects. Even if this information is not used during inference, it is still valuable.

Another open question is the feasibility of the up-block T2I. Whether or not it can be used to create detail is yet to be determined. After observing the loss function plateau, it was believed that the model had stopped learning. However, empirically, the images looked better at the end of training than they did at the beginning, despite similar loss values. This might be because the second dataset contained only close-ups, whereas the first had full-body images, leading to more masked areas and therefore higher loss. Thus, a higher loss might still indicate better quality.

To conclude, here are some areas that should be investigated further:
\begin{itemize}
    \item Testing the T2I adapter for up-blocks, and potentially increasing its parameter size.
    \item Utilizing pre-processing and labeling techniques for future models.
\end{itemize}

Additionally, improvements should be made in the following areas:
\begin{itemize}
    \item Using an IP adapter with a cross-attention dimension of 2028 as opposed to the 1280 used here. Multiple tokens can be injected as well.
    \item Ideally, the point-wise addition of a custom T2I adapter should occur in the cross-attention layers themselves, rather than between blocks.
    \item Measuring sleeve length as a continuous variable rather than using one-hot encoding.
\end{itemize}

\section{Other remarks}
The initial idea was to add an extra module to IDM-viton \cite{choi2024improvingdiffusionmodelsauthentic}, and add semantic information to that, however since they had trained in the same dataset as i had, it had already over-fitted to that dataset. Meaning that adding information about semantics is useless since the model already inputs things at the "correct" place. This lead me to have to pivot for another approach.

\begin{figure}[H]
    \centering
    \includegraphics[width=0.9\textwidth]{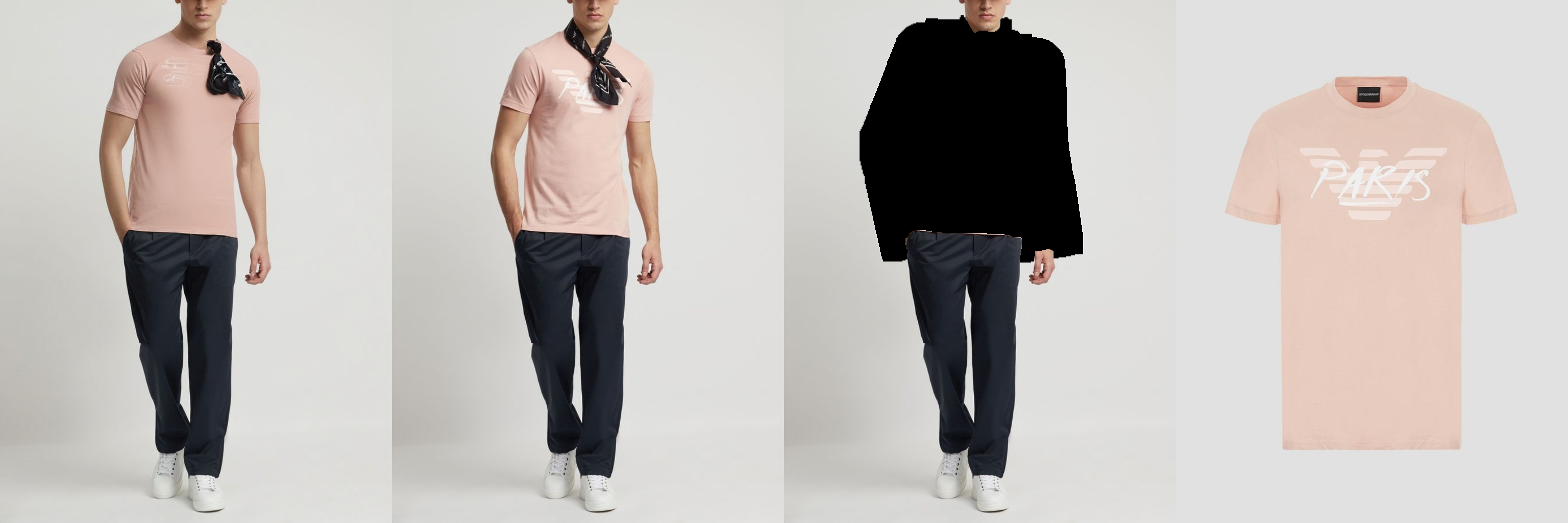}
    \caption{After a few epochs it became apparent that there was no more learning to be done as the model regenerated artefacts that should be hidden. Like this scarf on the left, that is completely covered by the mask.}
    \label{input}
\end{figure}

\newpage

\bibliographystyle{unsrt}
\bibliography{sources}

\appendix
\section{Appendix}

\subsection{Architechture of T2I}

\begin{table}[ht]
\centering
\begin{tabular}{|l|l|r|}
\hline
\textbf{Module} & \textbf{Sub-module} & \textbf{Parameters} \\ \hline
unshuffle & PixelUnshuffle & 0 \\ \hline
conv\_in1 & Conv2d & 2,458,112 \\ \hline
conv\_in2 & Conv2d & 5,899,520 \\ \hline
body & ModuleList & 72,105,280 \\ \hline
\multicolumn{2}{|l|}{AdapterBlock\_custom (0)} & 49,159,680 \\ \hline
 & AdapterResnetBlock (0) & 16,386,560 \\ \hline
 & \quad block1: Conv2d & 14,746,880 \\ \hline
 & \quad block2: Conv2d & 1,639,680 \\ \hline
\multicolumn{2}{|l|}{AdapterBlock\_custom (1)} & 19,666,560 \\ \hline
 & upsample: ConvTranspose2d & 6,554,880 \\ \hline
 & in\_conv: Conv2d & 819,840 \\ \hline
 & AdapterResnetBlock (0) & 4,097,280 \\ \hline
 & \quad block1: Conv2d & 3,687,040 \\ \hline
 & \quad block2: Conv2d & 410,240 \\ \hline
\multicolumn{2}{|l|}{AdapterBlock\_custom (2)} & 3,279,040 \\ \hline
 & in\_conv: Conv2d & 205,120 \\ \hline
 & AdapterResnetBlock (0) & 1,024,640 \\ \hline
 & \quad block1: Conv2d & 921,920 \\ \hline
 & \quad block2: Conv2d & 102,720 \\ \hline
\end{tabular}
\caption{Summary of selected modules and their parameters.}
\label{tab:modules}
\end{table}

\section{Generated Images}

\begin{figure}[H]
    \centering
    \includegraphics[width=0.99\textwidth]{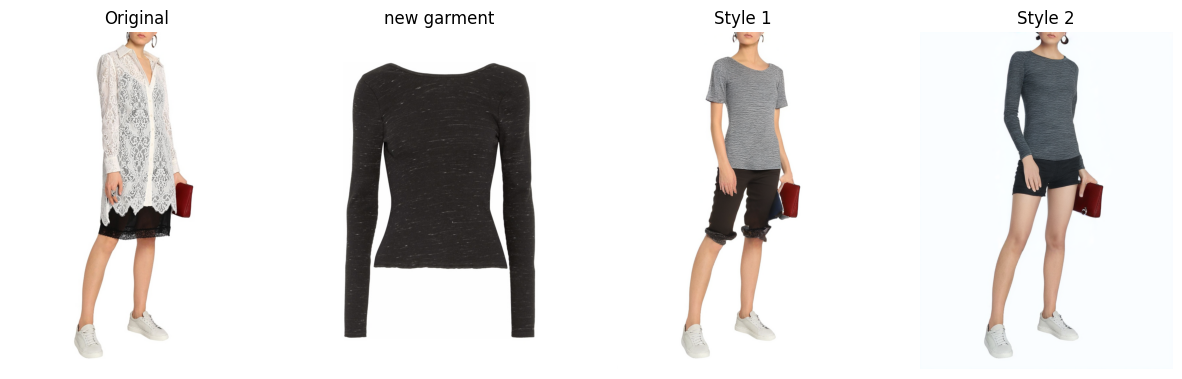}
    \caption{Different styles of outfitting a garment}
    \label{input}
\end{figure}

\begin{figure}[H]
    \centering
    \includegraphics[width=0.99\textwidth]{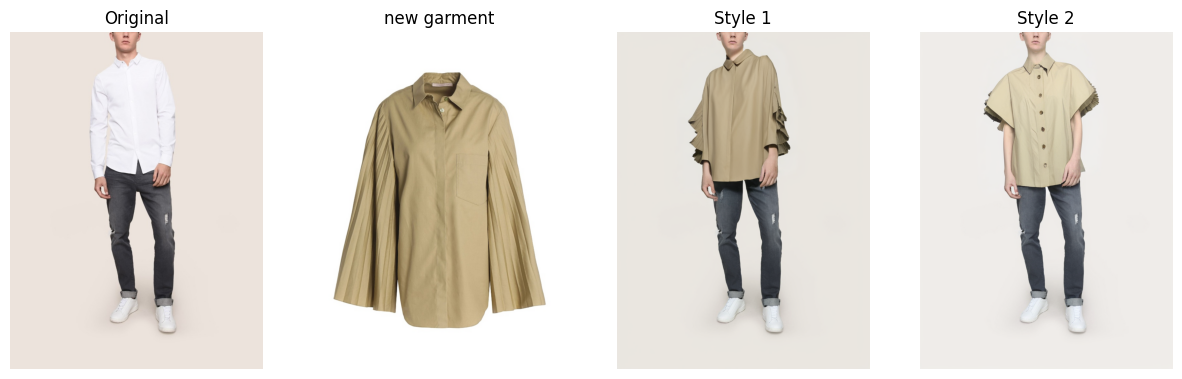}
    \caption{Different styles of outfitting a garment}
    \label{input}
\end{figure}

\begin{figure}[h!]
    \centering
    \subfigure{}
    \includegraphics[width=0.49\textwidth]{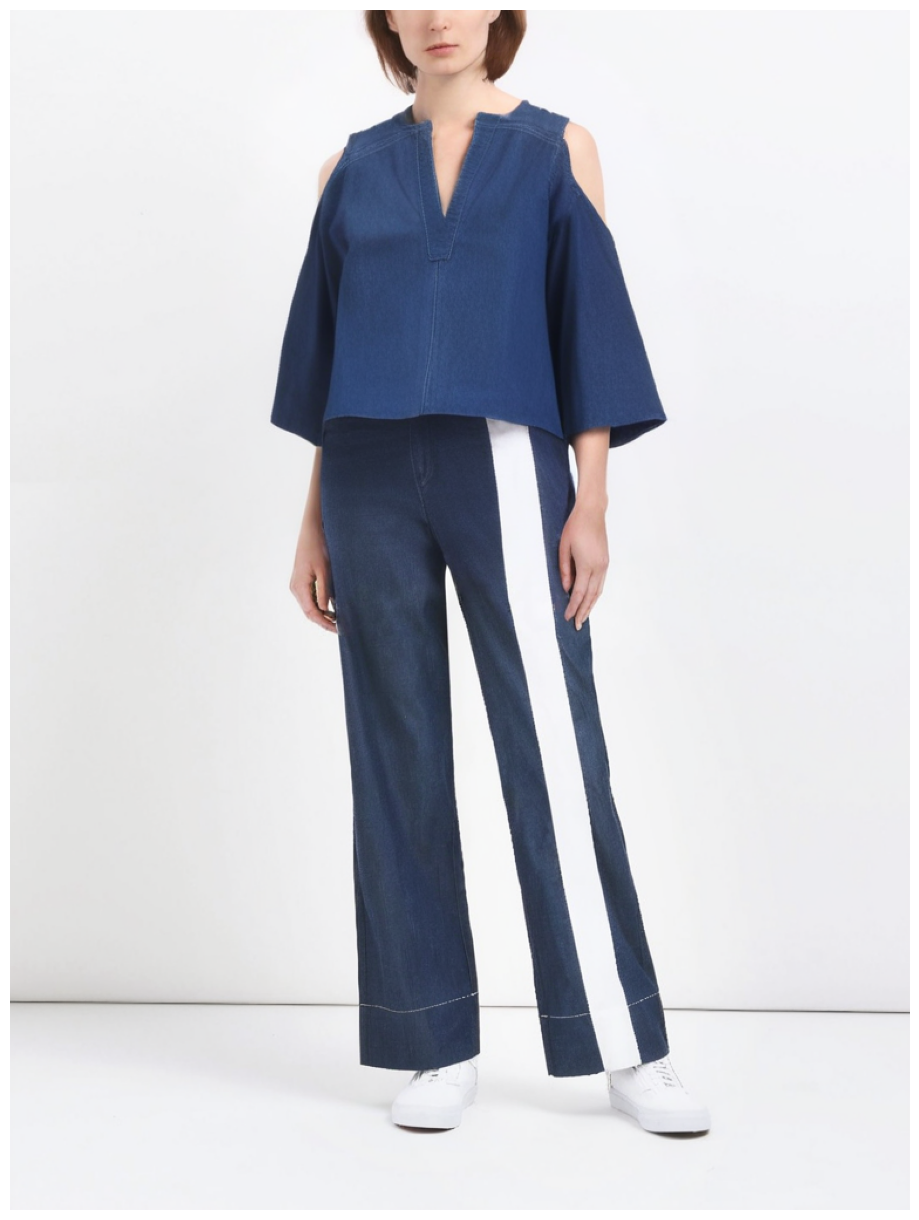}
    \subfigure{}
    \includegraphics[width=0.49\textwidth]{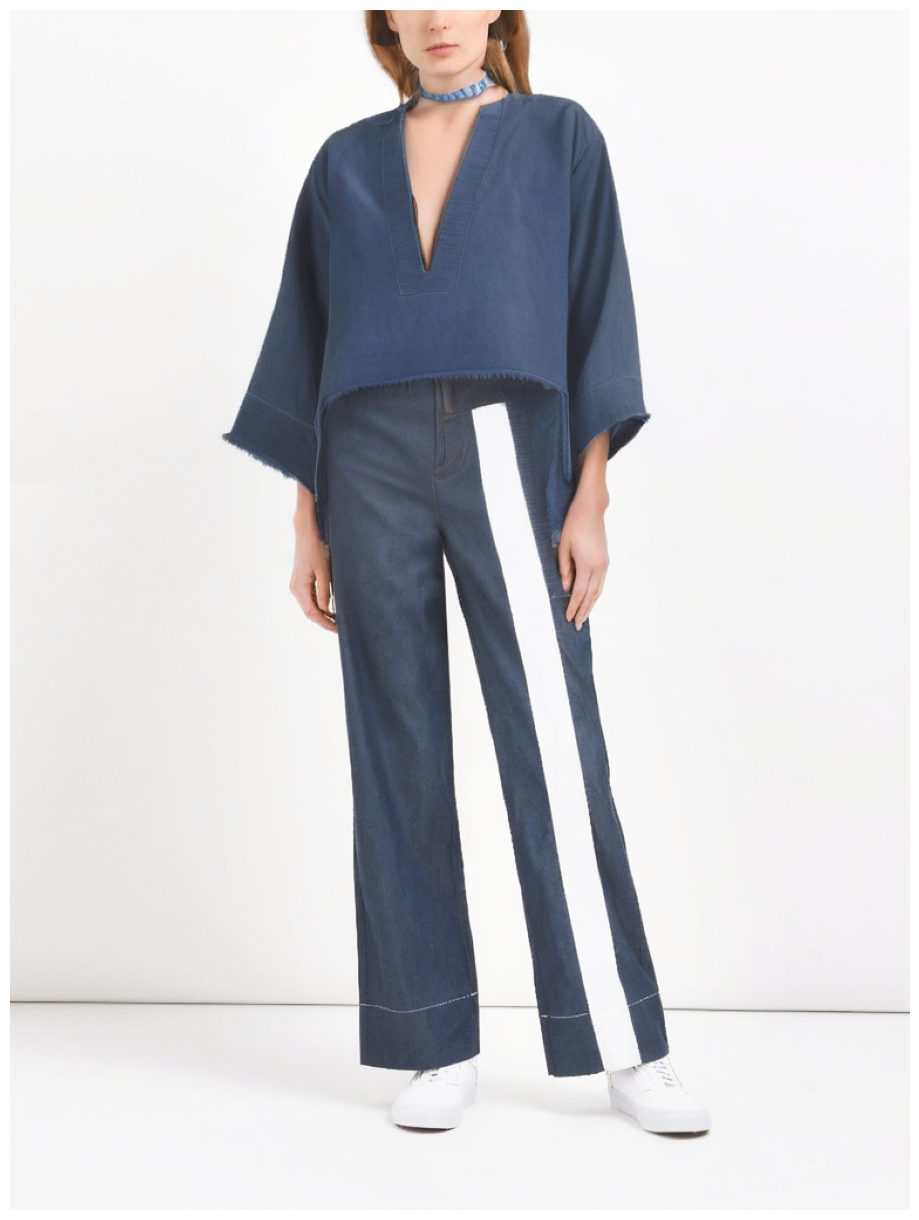}
    \caption{Further example of how changing shoulder-frac may change type of garment}
    \label{good}
\end{figure}

\end{document}